\documentclass[10pt,journal,compsoc]{IEEEtran}

\ifCLASSOPTIONcompsoc
  \usepackage[nocompress]{cite}
\else
  \usepackage{cite}
\fi
\ifCLASSOPTIONcompsoc
  \usepackage[caption=false,font=footnotesize,labelfont=sf,textfont=sf]{subfig}
\else
  \usepackage[caption=false,font=footnotesize]{subfig}
\fi
\usepackage{algorithm, algpseudocode}
\usepackage{amsmath, amssymb}
\usepackage{booktabs}
\usepackage{colortbl}
\usepackage{makecell, multirow}
\usepackage{mathtools}
\usepackage{rotating}
\usepackage{tabularx}
\usepackage{threeparttable, tablefootnote}
\usepackage{tasks}
\usepackage{url}
\usepackage{wasysym} 
\usepackage{xspace}
\usepackage{setspace}

\usepackage{ragged2e}

\usepackage[marginal]{footmisc}

\newcommand{\ie}{\emph{i.e.},\xspace}
\newcommand{\eg}{\emph{e.g.,}\xspace}

\newcommand{\etal}{\emph{et al.}\xspace}

\newcommand\figref[1]{Figure~\ref{#1}}

\newcommand\tabref[1]{Table~\ref{#1}}

\newcommand\secref[1]{$\S$~\ref{#1}}
\newcommand\equref[1]{Eq.(\ref{#1})}

\newcommand{\fakeparagraph}[1]{\vspace{1mm}\noindent\textbf{#1.}}

\newcommand{\systemname}{{\sf AdaDeep}\xspace}
\newcommand{\systemnameposs}{{\sf AdaDeep's}\xspace}

\renewcommand{\baselinestretch}{0.99}

\newcommand{\presec}{\vspace{0.0mm}}
\newcommand{\postsec}{\vspace{0.0mm}}

\newcommand{\lsc}[1]{{\color{black}#1}}
\newcommand{\lscrev}[1]{{\color{black}#1}}

\newcommand\zimu[1]{\textcolor{black}{#1}}

\ifodd 0
\newcommand{\rev}[1]{{\color{black}#1}} 
\else
\newcommand{\rev}[1]{#1}
\fi

\begin{document}

\title{\systemname: A Usage-Driven, Automated Deep Model Compression Framework for Enabling Ubiquitous Intelligent Mobiles}

\author{
Sicong Liu,~\IEEEmembership{Student~Member,~IEEE,}
Junzhao Du,~\IEEEmembership{Member,~IEEE},
Kaiming Nan, 
Zimu Zhou,~\IEEEmembership{Member,~IEEE},
Hui Liu,~\IEEEmembership{Member,~IEEE},
Atlas Wang,~\IEEEmembership{Member,~IEEE},
Yingyan Lin,~\IEEEmembership{Member,~IEEE}
\IEEEcompsocitemizethanks{
\IEEEcompsocthanksitem
Sicong Liu, Junzhao Du, Kaiming Nan, and Hui Liu are with the School of Computer Science and Technology, Xidian University, Xi'an, China;
Zimu Zhou is with School of Information Systems, Singapore Management University, Singapore;
Atlas Wang is with the Department of Computer Science and Engineering, Texas A$\&$M University, Texas, USA;
Yingyan Lin is with the Department of Electrical and Computer Engineering, Rice University, Texas, USA.
\protect\\
E-mail: \{Sicong Liu, Kaiming Nan\}@stu.xidian.edu.cn; \{dujz, liuhui\}@xidian.edu.cn; zimuzhou@smu.edu.sg; atlaswang@tamu.edu; yingyan.lin@rice.edu.
\IEEEcompsocthanksitem
Corresponding Author: Junzhao Du.
}
\thanks{Manuscript received May, 2019; revised December, 2019; accepted May, 2020.}}


\markboth{IEEE Transactions on Mobile Computing,~Vol.~0, No.~0, May~2020}%
{Liu \MakeLowercase{\textit{et al.}}: \systemname: A Usage-Driven Framework of On-Demand Deep Model Compression and Optimization for Mobile Devices}

\IEEEtitleabstractindextext{%
\justifying{
\begin{abstract}
Recent breakthroughs in Deep Neural Networks (DNNs) have fueled a tremendously growing demand for bringing DNN-powered intelligence into mobile platforms. While the potential of deploying DNNs on resource-constrained platforms has been demonstrated by DNN compression techniques, the current practice suffers from two limitations: 1) merely stand-alone compression schemes are investigated even though each compression technique only suit for certain types of DNN layers; and 2) mostly compression techniques are optimized for DNNs' inference accuracy, without explicitly considering other application-driven system performance (e.g., latency and energy cost) and the varying resource availability across platforms (e.g., storage and processing capability). To this end, we propose AdaDeep, a usage-driven, automated DNN compression framework for systematically exploring the desired trade-off between performance and resource constraints, from a holistic system level. Specifically, in a layer-wise manner, AdaDeep automatically selects the most suitable combination of compression techniques and the corresponding compression hyperparameters for a given DNN. 
Furthermore, AdaDeep also uncovers multiple novel combinations of compression techniques.
Thorough evaluations on six datasets and across twelve devices  demonstrate that \systemname can achieve up to $18.6\times$ latency reduction, $9.8\times$ energy-efficiency improvement, and $37.3\times$ storage reduction in DNNs while incurring negligible accuracy loss.
Furthermore, \systemname also uncovers multiple novel combinations of compression techniques.
\end{abstract}}}

\maketitle
\presec
\section{Introduction}
\IEEEPARstart{T}{here} is a growing trend to bring machine learning, especially deep neural networks (DNNs) powered intelligence to mobile devices~\lscrev{\cite{bib:arxiv2019:han}}.
Many smartphones and handheld devices are integrated with intelligent user interfaces and applications such as hand-input recognition (\eg iType\cite{bib:li2017:infocom}), speech-based assistants (\eg Siri), face recognition enabled phone-unlock (\eg FaceID).
New development frameworks targeted at mobile devices (\eg TensorFlow Lite) have been launched to encourage novel DNN-based mobile applications to offload the DNN inference to local mobile $\&$ embedded devices.
In addition to smartphones, DNNs are also expected to execute on-device inference on a wider range of mobile and IoT devices, such as wearables\cite{bib:yang2018:infocom} (\eg Fitbit wristbands) and smart home infrastructures (\eg Amazon Echo).
The diverse applications and the various mobile platforms raise a challenge for DNN developers and users:
\textit{How to adaptively generate DNNs for different resource-constrained mobile $\&$ embedded platforms to enable on-device DNN inference, while satisfying the domain-specific application performance requirements?}

Generating DNNs for mobile mobile $\&$ embedded platforms is non-trivial because many successful DNNs are computationally intensive while mobile $\&$ embedded devices are usually limited in computation, storage and power.
For example, LeNet~\cite{model:lenet}, a popular DNN for digit classification, involves 60k weight and 341k multiply-accumulate operations (MACs) per image. 
AlexNet~\cite{model:alexnet}, one of the most famous DNNs for image classification, requires 61M weights and 724M MACs to process a single image.
It can become prohibitive to download applications powered by those DNNs to local devices. 
These DNN-based applications also drain the battery easily if executed frequently.

In view of those challenges, \lscrev{DNN compression techniques have been widely investigated to enable the DNN deployment on mobile $\&$ embedded platforms by reducing the precision of weights and the number of operations during or after DNN training with desired accuracy. 
And consequently, they shrink the computation, storage, latency, and energy overhead on a target platform~\cite{bib:sze2017:arxiv},\lscrev{~\cite{ bib:arxiv2019:zhou}}.}
Various categories of DNN compression techniques have been studied, including weight compression~\cite{bib:bhattacharya2016:CD-ROM} \cite{bib:arXiv2015:han} \cite{bib:lane2016:IPSN} \cite{pmlr-v80-wu18h}, convolution decomposition~\cite{bib:ICLR2017:soravit} \cite{bib:arXiv2017:Howard} \cite{bib:cvpr2015:ICCV}, and special layer architectures~\cite{bib:iandola2016:arxiv} \cite{bib:lin2013:NIN}.
However, there are two major problems in existing DNN compression techniques:
\begin{itemize}
\item
Most DNN compression techniques aim to provide an \textit{one-for-all} solution without considering \lscrev{the diversity of} application performance requirements and  platform resource constraints.
\lscrev{
A single compression technique to reduce either model complexity or process latency may not suffice to meet complex user demands on the generated DNNs.
Both the selection of DNN compression techniques and the configuration of DNN compression hyperparameters should be \textit{on-demand}}, \ie adapt to the requirements and constraints on accuracy, computation, storage, latency, and energy imposed by developers and platforms.
\item
Most DNN compression techniques are \textit{manually} selected and \lscrev{configured through experience engineering, while the design criteria remain a black-box to non-expert end developers}.
An \textit{automatic} compression framework that allows user-defined criteria will benefit the development of DNN-powered mobile applications for diverse domain tasks.
\end{itemize}

This paper presents \systemname, a framework that automatically selects the compression techniques \lscrev{and the corresponding hyperparameters on a layer basis. It adapts to different user demands on application-specified performance requirements (\ie accuracy and latency) and platform-imposed resource constraints (\ie computation, storage, and energy budgets).
To integrate these complex user demands into \systemname, we formulate the tuning of DNN compression as a constrained hyperparameter optimization problem. 
}
\lscrev{
In particular, we define the DNN compression techniques (\eg weight compression and convolution decomposition techniques listed in \secref{subsec:bench_set}) as a new coarse-grained hyperparameter of DNNs. 
And we regard the compression hyperparameters (\eg the width multiplier and the sparsity coefficient enumerated in \secref{subsec:hyperparameter}) as the fine-grained hyperparameters of DNNs.
However, it is intractable to obtain a closed-form solution, due to 1) the large numbers of the coarse-grained hyperparameter, \ie combinations of DNN compression techniques, 2) the infinite search space of the fine-grained hyperparameters, \ie compression hyperparameters, and 3) the varying platform resource constraints.
Alternatively, \systemname applies a two-phase deep reinforcement learning (DRL) optimizer.
Specifically, it involves a deep Q-network (DQN) optimizer for compression technique selection, and a deep deterministic policy gradient (DDPG) optimizer for the corresponding compression hyperparameter search.
The two optimization phases are conducted interactively to provide a heuristic solution.
}

We implement \systemname with TensorFlow~\cite{url:tensorflow} and evaluate its performance over \lscrev{six} different public benchmark datasets for DNNs on twelve different mobile devices.
Evaluations show that \systemname enables a reduction of $1.7\times$ - $37.3\times$ in storage, $0.8\times$- $18.6\times$ in latency, $1.1\times$- $9.8\times$ in energy consumption, and $0.8\times$- $6.8\times$ in computational cost, with a negligible accuracy loss ($< 2.1\%$) for various datasets, tasks, and mobile platforms.

The main contributions of this work are as follows.
\begin{itemize}
\item
To the best of our knowledge, this is the first work that integrates the selection of both compression techniques and compression hyperparameters into an automated hyperparameter tuning framework, and balances the varied user demands on performance requirements and platform constraints.
\item
We propose a two-phase DRL optimizer to automatically select the best combination of DNN compression techniques \lsc{as well as the corresponding compression hyperparameters, in a layer-wise manner}.
\systemname extends the automation of DNN architecture tuning to DNN compression.
\item 
Experiments show that the DNNs generated by \systemname achieve much improved performance, as compared to existing compression techniques under various user demands (datasets, domain tasks, and target platforms).
\systemname also uncovers some novel combinations of DNN compression techniques suitable for mobile applications.
\end{itemize}

A preliminary version of \systemname has been published in \cite{bib:mobisys2018:liu}.
This work further develops \cite{bib:mobisys2018:liu} with the following three new contributions. 
First, a new DRL optimizer is proposed for fully automating the solving process of the constrained DNN compression problem (see Eq. (\ref{equ_opt})). Improving upon the one-agent based DQN for both the conv and fc layers, we develop a two-phase DRL optimizer for solving the constrained DNN compression problem in Eq. (\ref{equ_opt}). In particular, in the first phase \systemname leverages the separate DQN agents for conv and fc layers to select the optimal combination of compression techniques in a layer-wise manner (see \secref{sec:optimizer}), and then employs a DDPG optimizer in the second phase to search suitable compression hyperparameters for the selected compression techniques (refer to \secref{sec:hy_optimizer}). Second, all the experiments in \cite{bib:mobisys2018:liu} have been updated using the new DRL optimizer to extensively validate its effectiveness. Three, we have conducted experiments in additional model and dataset (\ie ResNet~\cite{model:resnet} on CIFAR-100~\cite{data:cifar100}) for evaluating \systemname in more diverse settings.

%
%

In the rest of this paper, we present \systemnameposs framework in \secref{sec:adadeep}, and formulate user demands on performance and resource cost in \secref{sec:model}. We present the overview of the automated two-phase DRL optimizer in \secref{sec:drl_overview}, and elaborate the design of these two types of optimizer in both \secref{sec:optimizer} and \secref{sec:hy_optimizer}. We evaluate \systemnameposs performance in \secref{sec:evaluation}, review the related work in \secref{sec:related}, \lscrev{discuss  limitations and future directions in \secref{sec:discuss},} and finally conclude this work in \secref{sec:conclusion}.
\vspace{-2mm}

\postsec
\presec
\section{Overview}
\label{sec:adadeep}

\begin{figure}[t]
  \centering
  \includegraphics[width=.42\textwidth]{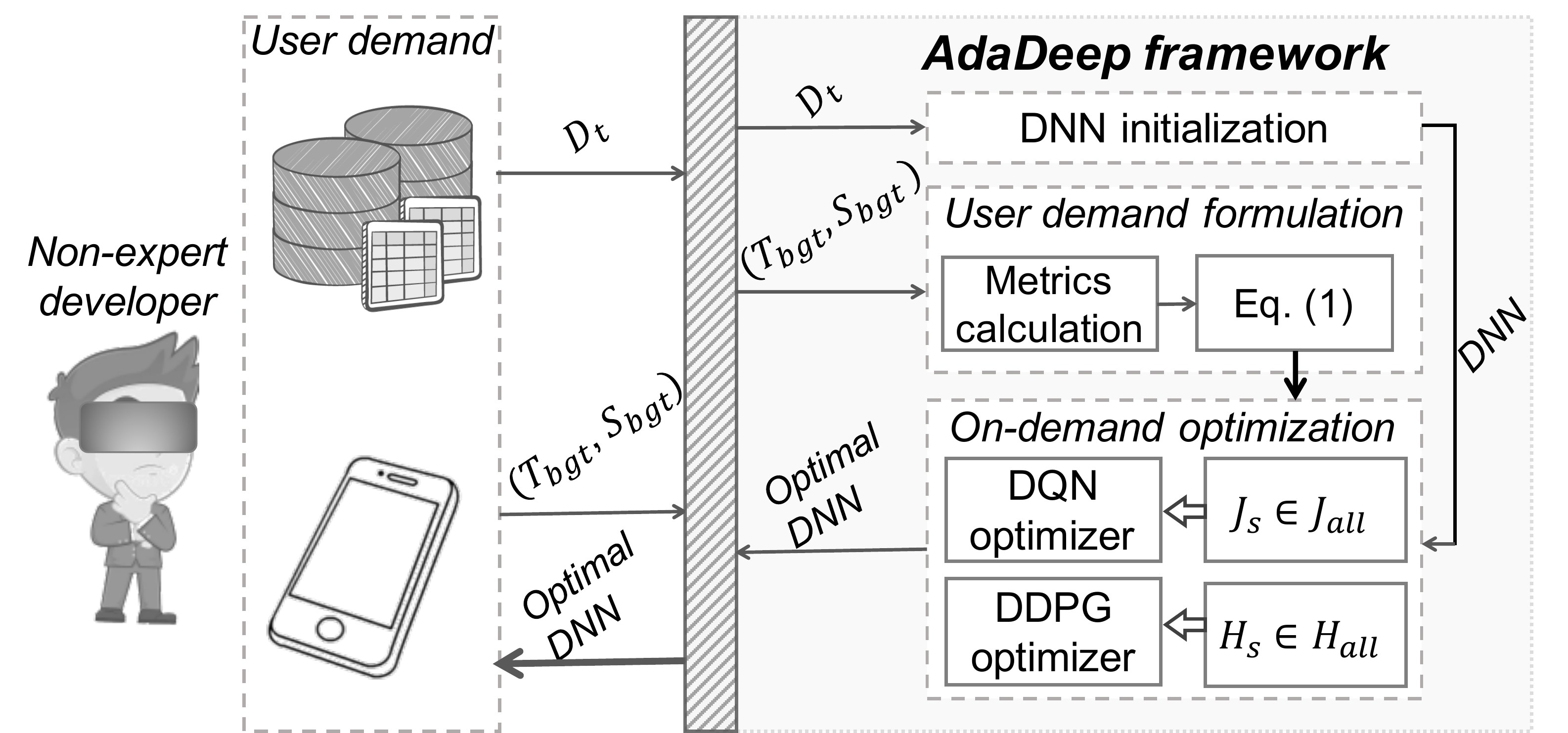}
  \vspace{-4mm}
  \caption{The block diagram of \systemname. \systemname accepts both system performance requirements and platform resource constraints from users (\eg DNN application developers), and then automatically generates a DNN that balances these requirements and constraints.}
  \label{fig:arc_adadeep}
\vspace{-4mm}
\end{figure}

This section presents an overview of \systemname.
From a system-level viewpoint, \systemname automatically generates the most suitable \textit{compressed} DNNs that meet the performance requirements and resource constraints imposed by end developers and the target deployment platforms.

\systemname consists of three functional blocks: \textit{DNN initialization}, \textit{user demand formulation}, and \textit{on-demand optimization} (\figref{fig:arc_adadeep}). 
The \textit{DNN initialization} block selects an initial DNN model for the \textit{on-demand optimization} block from a pool of state-of-the-art DNN models (\secref{subsec:setup}).
The \textit{user demand formulation} block quantifies the DNN's performance and cost (\secref{sec:model}), which are then input into the \textit{on-demand optimization} block as the optimization goals and constraints.
The \textit{on-demand optimization} block takes the initial DNN model and the optimization goals to automatically select the optimal DNN compression techniques \lsc{and compression hyperparameters} that maximize the system performance while satisfying cost budgets (\secref{sec:optimizer}).

Mathematically, \systemname aims to solve the following constrained optimization problem.
\begin{eqnarray}\label{equ_opt}
\mathop{arg max}\limits_{J_s \in J_{all}, H_s \in H_{all}} & \mu_1 \rev{N}(A-A_{min}) + \mu_2 \rev{N}(E_{max}-E) \nonumber \\
\text{s.t.} & T \leq T_{bgt}, \,\, S \leq S_{bgt},
\end{eqnarray}
where $A$, $E$, $T$ and $S$ denote the measured accuracy, energy cost, latency and storage of a given DNN running on a specific mobile platform.
User demands are expressed as a set of goals and constraints on $A$, $E$, $T$ and $S$.
Specifically, $A_{min}$ and $E_{max}$ are the minimal testing accuracy and maximal energy cost acceptable by the user.
The two goals on $A$ and $E$ are combined by importance coefficients $\mu_1$ and $\mu_2$.
$N(x)$ is a normalization operation, \ie $N(x)=(x-x_{min})/(x_{max}-x_{min})$.
We denote $T_{bgt}$ and $S_{bgt}$ as the user-specified latency and storage budgets.
The metrics $A$ and $S$ can be directly determined by the DNN architecture, while 
$E$ and $T$ are also platform-dependent.
However, all of them can be tuned by applying different DNN compression techniques and compression hyperparameters.
In summary, \systemname aims to select the best compression techniques $J_{s}$ from the set of all possible combinations $J_{all}$ \lsc{and search the optimal compression hyperparameter $H_s$ from the set of selective hyperparameter values $H_{all}$}, according to the user-demands on performance and resource budgets. 
For completeness, the set $J_{all}$ should be the permutations and combinations of discrete layer compression techniques at convolutional (conv) layers and fully-connected (fc) layers, defined as $J_{all}=Conv_{m_1}^{n_1} Fc_{m_2}^{n_2}$. 
Here, $m_1$ and $m_2$ are the number of optional compression techniques at conv and fc layers, respectively; $n_1$ and $n_2$ represent the number of conv and fc layers to be compressed, respectively; and the set $H_{all}$ is a continuous real-value space.
%

\rev{We maximize $A$, minimize $E$ while constrain $S$ and $T$ within the user-specified budgets, because we assume that accuracy is the most important performance metric, and the energy efficiency is in general more important than storage and latency for the power-sensitive mobile applications. 
\systemname can also integrate other optimization problem formulations.
}

Technically, \systemname faces two challenges.
\begin{itemize}
\item
It is non-trivial to derive the runtime performance $A$ and $S$, and the platform-dependent overhead $E$ and $T$ of a DNN. 
In \secref{sec:model}, \systemname proposes a systematic way to calculate these variables and associates them to the parameters of a DNN and the given  platform.
\rev{We apply the state-of-the-art estimation models and modify them to suite the software/hardware implementation considered in our work.
Evaluations show that the proposed estimation models can achieve the same ranking as the measured one on the real-world deployment platforms.}
\item
It is intractable to obtain a closed-form solution to the optimization problem in \equref{equ_opt}.
\systemname employs the deep reinforcement learning (DRL) based optimization process to solve it 
(see \secref{sec:drl_overview}, \secref{sec:optimizer}, and \secref{sec:hy_optimizer}).
Although DRL is a well-known optimization technique, its application in automated DNN architecture and hyperparameter optimization is emerging~\cite{zoph2016neural}.
We follow this trend and apply two types of layer-wise \zimu{DRL} optimizer, \ie deep Q-network (DQN) and deep deterministic policy gradient (DDPG), in the context of user-demand DNN compression. 
%
\end{itemize}

We summarize some symbols in \tabref{tb:symbol}, which are frequently used in this paper.

\begin{table}[t]
\caption{Summary of some frequently used symbols.}
\vspace{-2mm}
\scriptsize
\begin{tabular}{|c|c|}
\hline
\textbf{Symbols} & \textbf{Descriptions} \\ \hline
$A, E, T, C, S$ & \begin{tabular}[c]{@{}c@{}}DNN performance on accuracy,\\ energy, latency, computation, storage\end{tabular} \\ \hline
$T_bgt, S_bgt$ & latency and storage budgets \\ \hline
$\mu_1, \mu_2, \mu_3, \mu_4$ & \begin{tabular}[c]{@{}c@{}}Lagrange multiplier to balance \\ performance requirements and constraints \end{tabular} \\ \hline
\begin{tabular}[c]{@{}c@{}}$W_{1f}, W_{1c}, W_2, W_3, C_1$ \\ $C_2, C_3, L_1, L_2, L_3$ \end{tabular} & \begin{tabular}[c]{@{}c@{}}Ten mainstream DNN compression \\ techniques from three categories \end{tabular}
 \\ \hline
\end{tabular}
\vspace{-3mm}
\label{tb:symbol}
\end{table}

  \hfil

\postsec
\presec
\section{User Demand Formulation}
\label{sec:model}
This section describes how we formulate the user demand metrics, including accuracy $A$, energy cost $E$, latency $T$ and storage $S$, in terms of DNN parameters and platform resource constraints. 
Such a systematic formulation enables \systemname to predict the most suitable \textit{compressed} DNNs by user needs, \textit{before} being deployed to mobile devices.



\textbf{Accuracy $A$.}
The inference accuracy is defined as:
\begin{equation}
A = prob(\hat{d_i}= d_i), i \in D_{mb}
\end{equation}
where $\hat{d_i}$ and $d_i$ denote the classifier decision and the true label, respectively, and $D_{mb}$ stands for the sample set in the corresponding mini-batch.

\textbf{Storage $S$.} 
We calculate the storage of a DNN using the total number of bits associated with weights and activations \cite{pmlr-v70-sakr17a}:
\begin{equation}
S= S_f + S_p =\left|\mathcal{X}\right|B_{a}+\left|\mathcal{W}\right|B_{w}\label{eq: representational cost}
\end{equation}
where $S_f$ and $S_p$ denote the storage requirement for the activations and weights, $\mathcal{X}$ and $\mathcal{W}$ are the index sets of all activations and weights in the DNN. $B_{a}$ and $B_{w}$ denote the precision of activations and weights, respectively. 
For example, $B_{a} = B_{w} = 32$ bits in TensorFlow~\cite{url:tensorflow}. 

\textbf{Computational Cost $C$.} 
We model the computational cost $C$ of a DNN as the total number of  multiply-accumulate (MAC) operations in the DNN. 
For example, for a fixed-point convolution operation, the total number of MACs is a function of the weight and activation precision as well as the size of the involved weight and activation vectors \cite{RDSEC_SIPS}.

\textbf{Latency $T$.} 
The inference latency of a DNN executed in mobile devices strongly depends on the system architecture and memory hierarchy of the given device. 
We referred to the latency model in \cite{Latency} which has been verified in hardware implementations. 
Specifically, the latency $T$ is derived from a synchronous dataflow model, and is a function of the batch size, the storage and processing capability of the deployed device, as well as the complexity of the algorithms, \ie DNNs.



\textbf{Energy Consumption $E$.} 
The energy consumption of evaluating DNNs include computation cost $E_{c}$ and memory access cost $E_{m}$. 
The former can be formulated as the total energy cost of all the MACs, \ie $E_{c} = \varepsilon_{1} C$, where $\varepsilon_{1}$ and $C$ denote the energy cost per MAC operation and the total number of MACs, respectively. 
The latter depends on the storage scheme when executing DNNs on the given mobile device. 
We assume a memory scheme in which all the weights and activations are stored in a Cache and DRAM memory, respectively, as such a scheme has been shown to enable fast inference execution~\cite{bib:ISCA2016:chen} \cite{bib:arXiv2017:yang}\cite{bib:arxiv2017:xu}. 
Hence $E$ can be modeled as:
\begin{equation}
E=E_{c}+E_{m}= \varepsilon_{1} C + \varepsilon_{2} S_{p} + \varepsilon_{3} S_{f}
\end{equation}
where $\varepsilon_{2}$ and $\varepsilon_{3}$ denote the energy cost per bit when accessing the Cache and DRAM memory, respectively. 
To obtain the energy consumption, we refer to a energy model from a state-of-the-art hardware implementation of DNNs in~\cite{bib:arXiv2017:yang}, where the energy cost of accessing the Cache and DRAM memory normalized to that of a MAC operation is claimed to be $ 6$ and $200$, respectively. 
Accordingly:
\begin{equation}
E=\varepsilon_{1} \cdot   C + 6\cdot  \varepsilon_{1}\cdot  S_{p} + 200\cdot  \varepsilon_{1} \cdot S_{f}
\end{equation}
where $\varepsilon_{1}$ is measured to be $52.8$ pJ for mobile devices.

\textbf{Summary.} 
The user demand metrics ($A$, $S$, $T$ and $E$) can be formulated with parameters of DNNs (\eg the number of $C$, the index sets of all activations $\mathcal{X}$ and weights $\mathcal{W}$) and platform-dependent parameters (\eg the energy cost per bit).
The parameters of DNNs are tunable via DNN compression techniques and compression hyperparameters.
Different mobile platforms vary in platform parameters and resource constraints.
Hence it is desirable to \textit{automatically} select appropriate compression techniques and compression hyperparameters to optimize the performance  and resource cost for each application and platform.

Note that it is difficult to precisely model the platform-correlated user demand metrics, \eg $E$ and $T$, since they are tightly coupled with the platform diversity.
However, the ranking of the DNNs costs derived by the above estimation models is consistent with the ranking of the actual costs of these DNNs measured on the real-world deployment platforms. 
As will be introduced in the next section, the proposed \systemname framework is generic and it can easily integrate other advanced estimation models.

\postsec
\presec
\section{On-demand Optimization Using DRL}
\label{sec:drl_overview}
We leverage deep reinforcement learning (DRL) to solve the optimization problem in \equref{equ_opt}. 
\lsc{
Specifically, two types of DRL optimizers are employed to automatically select compression techniques and the corresponding hyperparameters (\eg compression ratio, number of inserted neurons, and sparsity multiplier) on a layer basis, in the goal of maximizing performance requirements (\ie $A$ and $E$) while satisfying users' demands on cost constraints (\ie $S$ and $T$).}

\begin{figure}[t]
  \centering
  \includegraphics[width=0.47\textwidth]{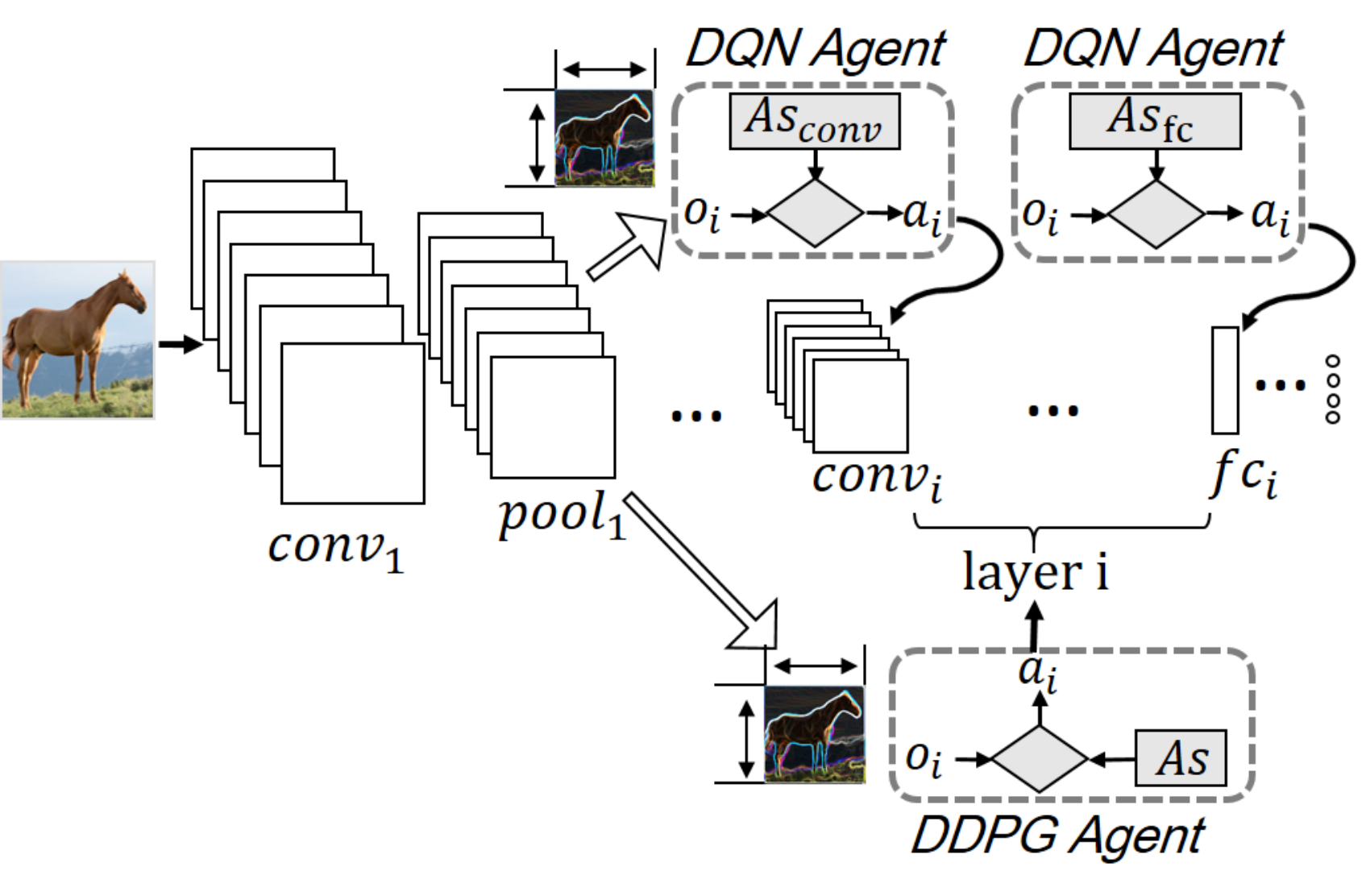}
  \vspace{-4mm}
  \caption{The proposed two-phase DRL optimization
  for \equref{equ_opt}. 
  It takes performance requirements and cost constraints as its input, automatically selects compression techniques and the corresponding compression hyperparameters in a layer-wise manner by using the DQN and DDPG agents, and outputs an optimally compressed DNN.
  }
   \vspace{-4mm}
  \label{fig_alc}
\end{figure}

\lsc{
\figref{fig_alc} shows the two-phase DRL optimizer designed for the automated DNN compression problem.
The first phase leverages two DQN agents for conv and fc layers to select a suitable combination of \textit{compression techniques} in a layer-wise manner.
A DDPG optimizer agent is used in the second phase to select \textit{compression hyperparameter} from a continuous real-value space for the selected compression techniques at different layers.
The two optimization phases are conducted interactively.
During the DQN-based optimization phase, the hyperparameters at different compressed layers are fixed as the values estimated by DDPG agent.
In the DDPG-based optimization phase, hyperparameter search is performed based on the compression techniques selected by DQN.}

DQN and DDPG are two typical DRL methods to handle complex input, action and rewards to learn the controller agent.
In the literature of DRL, a \textit{policy} $\pi$ refers to a specific mapping from  \textit{state} $o$ to \textit{action} $a$. 
A reward function $R(o,o',s)$ returns the gain when transitioning to state $o'$ after taking action $a$ in state $o$.
Given a state $o$, an action $a$ and a policy $\pi$, the action-value (a.k.a. the $Q$ function) of the pair ($o$, $a$) under $\pi$ is defined by the \textit{action-value}, which defines the expected reward for taking action $a$ in state $o$ and then following policy $\pi$ thereafter.
The DQN agent iteratively improves its $Q$-function by taking actions, observing the reward and next state in the environment, and updating the estimate.
Once the DQN agent is learned, the optimal policy for each state $o$ can be decided by selecting $a$ with the highest $Q$-value.
\lsc{As for the DDPG agent, it involves an actor-critic framework to combine the idea of DQN and Policy Gradient.
Policy Gradient seeks to optimize the policy space directly, that is, an actor network learns the deterministic policy $\pi$ to select action $a$ at state $o$. 
And a value-based critic network is to evaluate potential value $Q$ of policy $(o, a)$ estimated by the actor network.
}
%
We propose to adopt the DRL, \lsc{\ie DQN and DDPG}, for automated DNN compression in \systemname for the following reasons:
\begin{itemize}
\item
\lsc{Both DQN and DDPG agents} enable  automatic decision based on the dynamically detected performance and cost. And they are suited for non-linear and non-differentiable optimization.
\item
The \lsc{DNN to be compressed} and the DQN or DDPG agent can be trained jointly end-to-end~\cite{bib:acc2017:liu}.
\lsc{Because the DQN/DDPG engent} employs the neural network architecture, therefore can participant the feed-forward and back-propagation operations of the DNNs to be compressed.
And the output of DQN and DDPG is the decision signal to control the selection of compressed techniques and hyperparameters.
\item
The DRL-based optimizer provide both capability and flexibility in DNN compression.
Within the framework of DQN, we can easily add or delete selective compression techniques by simply adding branch sub-networks (\ie actions), and figure out the mapping function of the complex optimization problem's input and results.
\lsc{And DDPG can also expand or narrow the value region (action space) without affecting other components of this framework.}
\end{itemize}

To apply DRL to the DNN compression problem, we need to 
\textit{(i)}
design the reward function to estimate the immediate reward and future reward after taking an action;
\textit{(ii)}
design the definition of DRL's state $o$ and action $a$ in the context of DNN compression; and
\textit{(iii)}
design the DRL architecture and training algorithm with tractable computation complexity.
We will elaborate them in \secref{sec:optimizer} and \secref{sec:hy_optimizer}.
We note that the proposed two-phase DRL optimizer, \ie DQN- and DDPG-based optimizer, are still heuristic.
Hence they cannot theoretically guarantee a globally optimal solution.
However, as we will show in the evaluations, the proposed optimizer outperform exhaustive or greedy approaches in terms of the performance of the compressed DNNs.

\postsec
\presec
\section{DQN Optimizer for Layer-wise Compression Technique Selection}
\label{sec:optimizer}

\begin{table*}[ht]
\scriptsize
\centering
\caption{The DQN terms explained in the context of DNN compression technique selection.}
\label{tb_semantic}
\vspace{-4mm}
\begin{tabular}{|c|c|}
\hline
\textbf{DQN Terms} & \textbf{Contextual Meanings for DNN compression} \\ \hline
State $o_i$$\sim$$O$s & Input feature size to DNN layer $i$ \\ \hline
Action $a_i$$\sim$$A$s & Selective compression techniques for DNN layer $i$ \\ \hline
Reward function $R$ & Optimization gain $G$ \& constraints satisfaction $H$ \\ \hline
$Q$ value = $\gamma \sum R$ & Potential optimization gain \& constraints satisfaction \\ \hline
Training loss function & Difference between the true $Q$ value and the estimated $Q$ value of DQN \\ \hline
\end{tabular}
\vspace{-3mm}
\end{table*}

\subsection{Design of Reward Function}
\label{subsec:reward}
To define the reward function $R$ according to the optimization problem \equref{equ_opt}, a common approach is to use the Lagrangian Multiplier~\cite{bib:SIAM2008:Ito} to  convert the constrained formulation into an unconstrained one:
\begin{equation}
\label{eq_multiplier1}
\begin{split}
& \ R=\ [\mu_1 Norm(A-A_{min}) + \mu_2 Norm (E_{max}-E) \\
&  + \mu_3 Norm(T_{bgt}-\frac{C}{P})  + \mu_4 Norm(S_{Cache}-S_p)] \\
\end{split}
\end{equation}
where $\mu_1$, $\mu_2$, $\mu_3$ and $\mu_4$ are the Lagrangian multipliers. 
It merges the objective (\eg $A$ and $E$) and the constraint satisfaction (how well the $T$ and $E$ usages meet budgets). 
However, maximizing \equref{eq_multiplier1} rather than \equref{equ_opt} will cause ambiguity. For example, the following two situations lead to the same objective values are thus indistinguishable: \textit{(i)} poor accuracy and energy performance, with low latency/storage usage; and \textit{(ii)} high accuracy and energy performance, with high latency/storage usage. 
Such ambiguity can easily result in a compressed DNN that exceeds the user-specified latency/storage budgets.

\begin{figure}[t]
  \centering
  \includegraphics[width=0.47\textwidth]{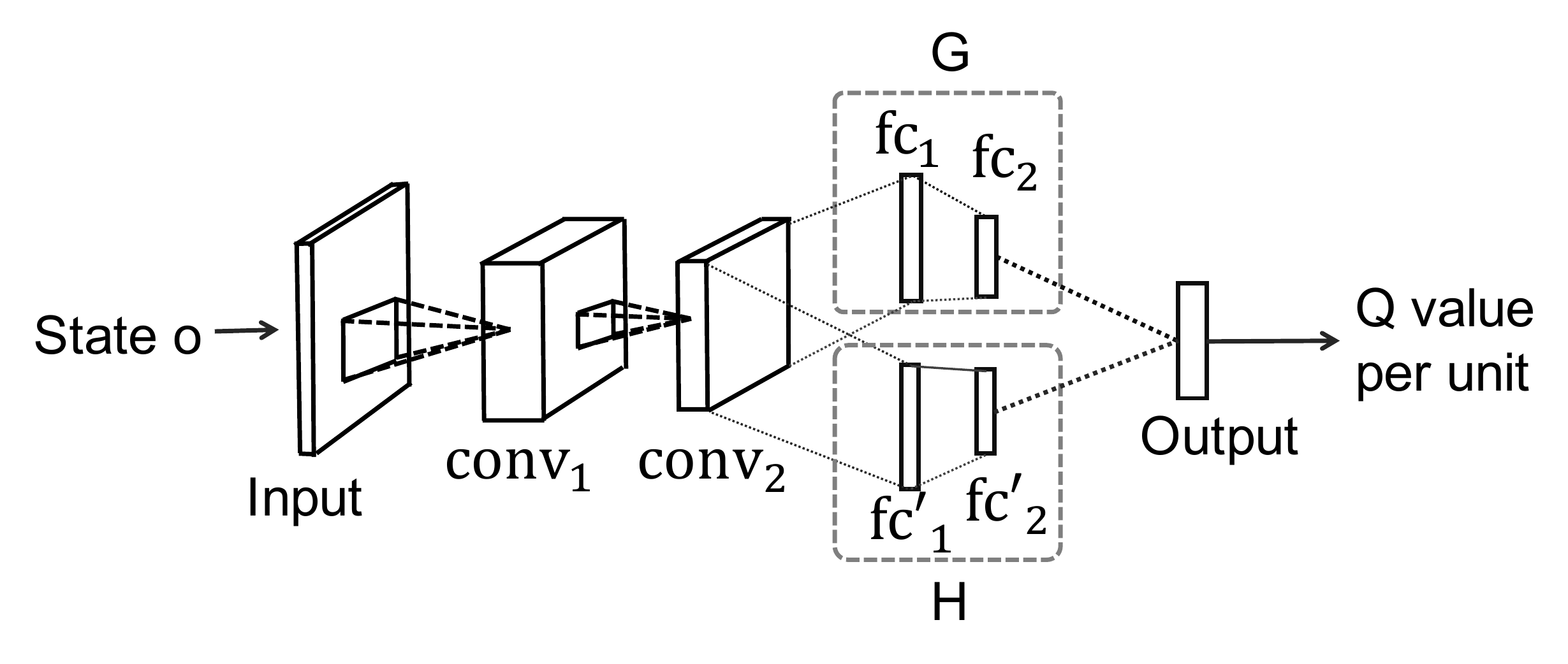}
  \vspace{-2mm}
  \caption{The dueling DQN architecture for  conv and fc layers.}
  \vspace{-3mm}
\label{fig_dqn}
\end{figure}

To avoid such ambiguity, we define two loss functions for the objective gain and the constraint satisfaction, respectively.
We borrow the idea of dueling DQN~\cite{bib:wang2016:arxiv} to separate the state-action value function and the state-action advantage function into two parallel streams (see \figref{fig_dqn}). 
The two streams share conv layers with parameters $\omega$ which learn the \rev{representations} of states. And then they joint two columns to separately generate the state-action objective gain value $G$, with weight parameter $\beta$, and the state-action constraint satisfaction value $H$, with weight parameter $\eta$.
The two columns are finally aggregated to output a single state-action value $Q$.
We define a novel $Q$ value:
\begin{equation}
Q(o,a;\omega,\beta,\eta)=G(o,a;\omega,\beta)+H(o,a;\omega,\eta)	
\end{equation}
The network $G$ and $H$ comes with their corresponding reward functions $R_1$ and $R_2$:
\begin{equation}
\label{equ_DRL_R2}
\begin{split}
& R_1 = \mu_1  Norm(A-A_{min}) - \mu_2 Norm (E_{max}-E) \\
& R_2 = \mu_3 Norm(T_{bgt}-\frac{C}{P}) + \mu_4 Norm(S_{Cache}-S_p) \\
\end{split}
\end{equation}
After taking an action, we observe the rewards $R_1$ for $G$ and $R_2$ for $H$, and use their interaction and balance to guide the selection of compression techniques.


\begin{algorithm}[ht]
\caption{DQN optimizer for Layer-wise Compression Technique Selection and Combination} \label{alg_drl_alc}
\begin{algorithmic}[1]
\Procedure {DQN}{$D_t$, $Budgets$, $As\_conv$, $As\_fc$}
\State Initialize $\Lambda$, $DNN$
\State Initialize DQN's predict $Q$ with random $\omega, \beta, \eta $
\State Initialize DQN's target $\overline{Q}$ with weights $\overline{\omega}, \overline{\beta}, \overline{\eta}$
\For{episode in range(1000)}
    \While{layer $i$ is to be compressed}
        \State observe state $o_i$ at layer $i$
        \If{$i$ is conv layer}
            \State select $a_i$ from $As\_conv$ at $o_i$ by $Q$ value ($\epsilon-greedy$)
        \ElsIf{$i$ is fc layer}
            \State select $a_i$ from $As\_fc$ at $o_i$ by $Q$ value ($\epsilon-greedy$)
        \EndIf
        \State layer $i++$
    \EndWhile
    \State forward DNN to compute Reward $R_{1t}, R_{2t}$
    \State broadcast $R_t$ to be the reward of all $o_i$
    \State store transmission $(o_i,a_i,R_t,o_{i+1})$ of each layer in reply memory $\Lambda$ 
    \State $\overline{Q}_t= R_{1t}+R_{2t}+\gamma Q(o', argmax Q(o',a'; \omega_{i},\beta_{i},\eta_{i});\overline{\omega},\overline{\beta},\overline{\eta})$
    \State perform greedy descent to update DQN's $\omega$ on loss of random mini-batches replay: 
    \State $L(\omega)=\mathbb{E}_{(o,a,R,o') \sim \Lambda} (\overline{Q_t}- Q(o_i,a_i; \omega_i, \beta_i, \eta_i))^2$ 
    \State every $num$ steps reset $\overline{Q}=Q$
\EndFor
\EndProcedure
\end{algorithmic}
\end{algorithm}

\subsection{The DQN Optimizer for Compression Technique Selection and Combination}

The proposed layer-wise DQN optimizer for compression technique selection and combination is outlined in Algorithm \ref{alg_drl_alc}.
\tabref{tb_semantic} explains the contextual definitions of the DQN terms in our compression technique selection problem.
\lsc{
For each layer $i$, we observe a state $o_i$.
%
Two agents are employed for two types of DNN layers (\ie conv and fc layers), which respectively regard the optional compression techniques at conv and fc layers as their action space $As_{conv}$ and $As_{fc}$. 
For each layer/state $o_i$, we select a random action with probability $\epsilon$ and select the action with largest $Q$ value by $1-\epsilon$ probability ($\epsilon = 0.001$ by default).
Repeating the above operation layer by layer, we forward the entire DNN to compute a global Reward $R_t =R_{t1}+R_{t2}$, and regard it as the reward of each states $o_i$.
}

To build a DQN with weight parameters $\omega$, $\beta$ and $\eta$, we optimize the following loss function iteratively. At iteration $t$, we update  $Q(o,a;\omega_t,\beta_t,\eta_t)$.
\begin{equation}
L(\omega_t)= \mathbb{E}_{(o,a,R,o')\sim \Lambda}[{(\overline{Q}_{i}-Q(o,a;\omega_t,\beta_t,\eta_t))}^2]
\end{equation}
with the frozen $Q$ value learned by the target network~\cite{bib:van2016:AAAI}:
\begin{equation}
\overline{Q}_{t}= R_{1t}+R_{2t}+\gamma Q(o', \max Q(o',a'; \omega_t, \beta_t, \eta_t);\overline{\omega},\overline{\beta},\overline{\eta})
\end{equation}
We adopt the standard DQN training techniques~\cite{bib:wang2016:arxiv} and use the update rule of SARSA~\cite{bib:ADPRL2009:van} with the assumption that future rewards are discounted by a factor $\gamma$ \cite{bib:mnih2013:arxiv} of the default value $0.01$. 
And we leverage experience replay to randomly sample from a memory $\Lambda$, to increase the efficiency of DQN training.

\postsec
\presec

\lsc{\section{The DDPG Optimizer for Compression Hyperparameter Search}
\label{sec:hy_optimizer}

We employ a DDPG optimizer to automatically search the proper compression hyperparameters for layer compression techniques from a continuous action space~\cite{bib:eccv2018:he}. 
Its contextual definitions of state $o$ and reward $R$ are the same as that in the DQN optimizer~(see \tabref{tb_semantic}).

\begin{figure}[ht]
  \centering
  \includegraphics[width=0.4\textwidth]{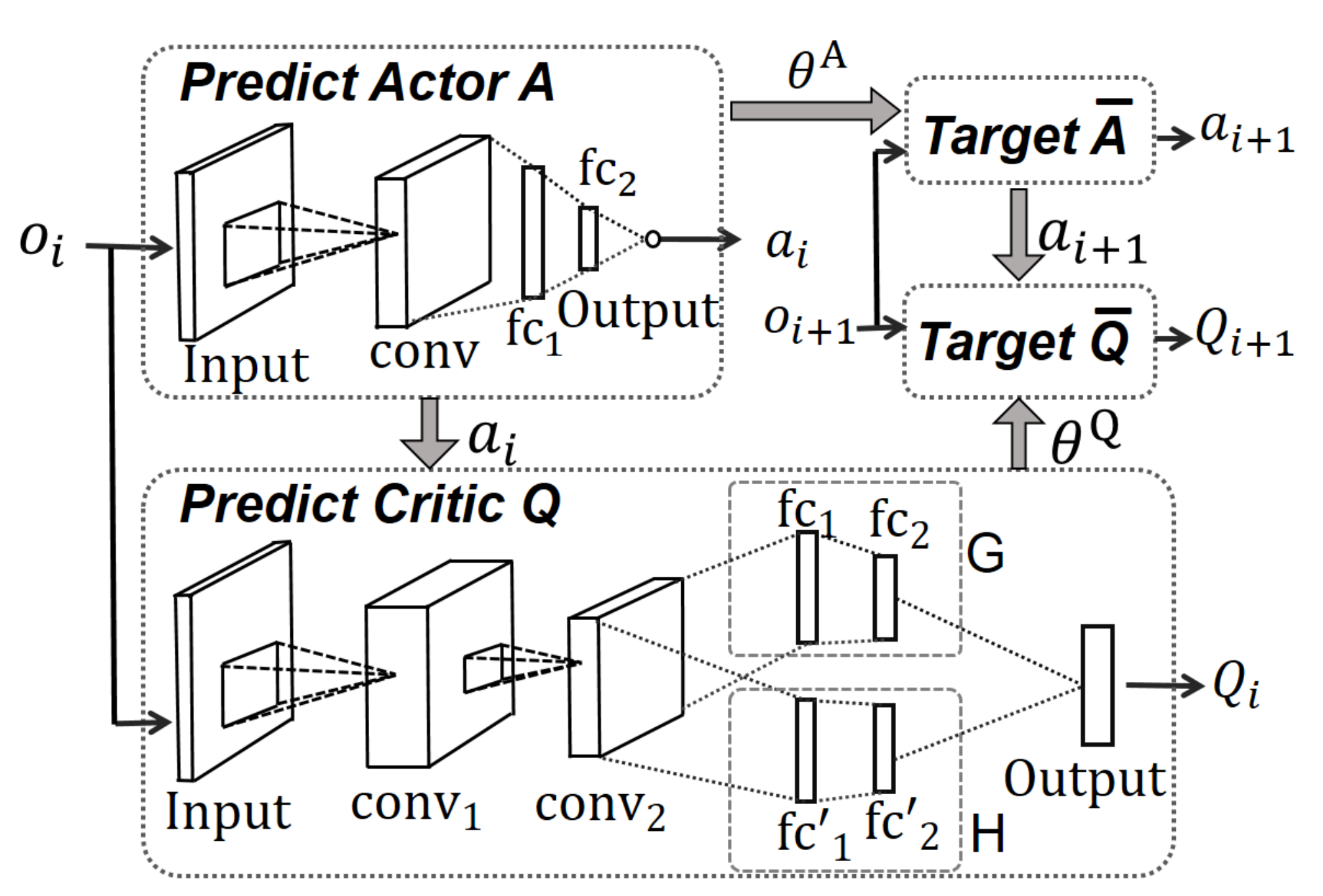}
  \vspace{-2mm}
  \caption{The architecture of the DDPG optimizer for each layer/state $o_i$.}
\label{fig_ddpg}
\end{figure}

\textbf{Action Space $A_s$ for Hyperparameter Search.}
The compression hyperparameters considered in this work include the compression ratio in a weight pruning~\cite{bib:arXiv2015:han}, the number of inserted neurons by weight factorization~\cite{bib:lane2016:IPSN, bib:bhattacharya2016:CD-ROM}, and the sparsity multiplier in a convolution decomposition~\cite{bib:arXiv2017:Howard, bib:ICLR2017:soravit}. 
Note that we search hyperparameters from a continuous action space for its effectiveness.
To simplify implementation and reduce the training time, we transfer all of the above compression hyperparameters into a ``ratio'', whose value space is mapped into $[0,1]$, so that we only need one DDPG agent to select action $a$ from the same action space $A_s \sim [0,1]$ for all compressed layers.
We defer the transformation details from compression hyperparameters to the ratio to \secref{subsubsec:performance_optimizer}.

\figref{fig_ddpg} shows the architecture of the proposed DDPG optimizer. 
It follows a actor-critic framework to concurrently learn the actor network $A$ and the value-based critic network $Q$.
The actor $A$ gets advice from the critic $Q$ that helps the actor decide which actions to reinforce during training.
Meanwhile, the DDPG makes uses of double actor networks and critic networks to improve the stability and efficiency of  training~\cite{bib:ICLM2014:silver}.
The architecture of $\overline{A}$ and $\overline{Q}$ is the same as $A$ and $Q$ with frozen parameters.
We adopt a same dueling DQN architecture (see \figref{fig_dqn}) to build the critic network $Q$ and $\overline{Q}$, which separates the reward into objective gain $G$ and constraint satisfaction $H$ (refer to \secref{subsec:reward}).
And we establish the actor network, expressing the deterministic state-action function, through several conv and fc layers with parameters $\theta^A$.

Algorithm \ref{alg_ddpg} illustrates the DDPG optimizer for compression hyperparameter search.
For each compressed layer $i$, it observes a state $o_i$ and leverages the DDPG's predict actor network $A$ to estimate the deterministic optimal action $a_i$ with truncated normal distribution noise $\epsilon$~\cite{bib:eccv2018:he}.
Repeating above operations, it forwards the DNN network to compute a global reward $R_{1t}$ and $R_{2t}$, which is broadcast to each layer/state $o_i$.
Then the predict critic network $Q$ estimates the state value $Q_i$ of the current state $o_i$ and of the action $a_i$ estimated by the actor $A$.

\begin{algorithm}[t]
\caption{DDPG optimizer for Compression Hyperparameter Search} \label{alg_ddpg}
\begin{algorithmic}[1]
\Procedure {DDPG}{$D_t$, $Budgets$, $As$}
\State Initialize $\Lambda$, $DNN_{compressed}$
\State Initialize predict actor network $A(\theta^A)$ and target $\overline{A}(\overline{\theta^A})$ 
\State Initialize predict critic network $Q(\theta^Q)$ and target $\overline{Q}(\overline{\theta^Q})$ 
\For{episode in range(1000)}
    \While{layer $i$ is to be compressed}
        \State observe state $o_i$ at layer $i$
        \State select $a_i$ from $As$ at $o_i$ by $A$ with noise $\epsilon$
        \State layer $i++$
    \EndWhile
    \State forward DNN network to compute Reward $R_{1t}, R_{2t}$
    \State broadcast $R_t$ to be the reward of all $o_i$
    \State store transmission $(o_t,a_t,R_{1t},R_{2t},o_{t+1})$ of all states in $\Lambda$
    \State update $A$ using policy gradient:
        \State $\bigtriangledown_{\theta^A}\approx \frac{1}{N}\sum_{t}{\bigtriangledown_a}Q(o_t,A(o_t);\theta^Q) \bigtriangledown_{\theta^A} A(o_t;\theta^A)  $ 
        \State set $y_t = R_{1t}+ R_{2t} + \gamma \overline{Q}(o_{t+1}, \overline{A}(o_{t+1};\overline{\theta ^A}))$ 
        \State update $Q$ on loss of random mini-batches replay: $L=\frac{1}{A}\sum_{t}( {y_t- Q(o_t, a_t; \theta^Q)})^2$
   \State every $num$ steps reset $\overline{Q}=Q$ and $\overline{A}=A$\;
\EndFor
\EndProcedure
\end{algorithmic}
\end{algorithm}

To train such DDPG optimizer, we optimize the actor network $A$ at iteration $t$ via the policy gradient function:
\begin{equation}
\label{loss_A}
\bigtriangledown_{\theta^A}\approx \frac{1}{N}\sum_{t}{\bigtriangledown_a}Q(o_t,A(o_t);\theta^Q) \bigtriangledown_{\theta^A} A(o_t;\theta^A)  
\end{equation}
And we train the critic network $Q$ by optimizing the loss function $L$ from both the random reply memory and the output of the actor and the critic networks:
\begin{equation}
\label{eq_multiplier}
\begin{split}
y_t = R_{1t}+R_{2t}+ \gamma \overline{Q}(o_{t+1}, \overline{A}(o_{t+1};\overline{\theta^A}))  \\
L=\frac{1}{A}\sum_{t}( {y_t- Q(o_t, a_t; \theta^Q)})^2
\end{split}
\end{equation}
where $y_t$ is computed by the sum of immediate reward $R_{1t}$ and $R_{2t}$ and the outputs of the frozen actor $\overline{A}$ and critic $\overline{Q}$.

}
\postsec
\presec
\section{Evaluation}
\label{sec:evaluation}
This section presents evaluations of \systemname across various mobile applications and platforms.

\subsection{Experiment Setup}
\label{subsec:setup}
We first present the settings for our evaluation.

\textbf{Implementation.}
We implement \systemname with TensorFlow~\cite{url:tensorflow} in Python. %
The \textit{compressed} DNNs generated by \systemname are then loaded into the target platforms and evaluated as Android projects executed in Java. 
Specifically, \systemname selects an initial DNN architecture from a pool of three state-of-the-art DNN models, including LeNet~\cite{model:lenet}, AlexNet~\cite{model:alexnet}, ResNet~\cite{model:resnet}, and VGG~\cite{model:vgg}, according to the size of samples in $D_t$. 
For example, LeNet is selected when the sample size is smaller than $28\times28$, otherwise AlexNet, VGG, or ResNet is chosen. 
Standard training techniques, such as \rev{stochastic} gradient descent (SGD) and Adam~\cite{bib:arxiv2014:kingma}, are used to obtain weights for the DNNs.

\textbf{Evaluation applications and DNN configurations.}
To evaluate \systemname, we consider six commonly used mobile tasks.
Specifically, \systemname is evaluated for hand-written digit recognition ($D_{1}$: MNIST ~\cite{data:mnist1998:LeCun}), image classification ($D_{2}$: CIFAR-10~\cite{data:cifar}, \lscrev{ $D_{3}$: CIFAR-100~\cite{data:cifar100}} and $D_{4}$: ImageNet~\cite{data:imagenet}), audio sensing application ($D_{5}$: UbiSound~\cite{bib:sicong2017:IMWUT}), and human activity recognition ($D_{6}$: Har~\cite{data:Har}). According to the sample size, LeNet \cite{model:lenet} is selected as the initial DNN structure for $D_{1}, D_{2}$, $D_{5}$ and $D_{6}$, \lscrev{ResNet-56 is choosen for $D_{3}$}, while AlexNet \cite{model:alexnet} and VGG-16 \cite{model:vgg} are chosen for $D_{4}$. 


\textbf{Mobile platforms for evaluation.}
We evaluate \systemname on \textbf{twelve commonly used mobile and embedded platforms}, including six smartphones, two wearable devices, two development boards and two smart home devices, which are equipped with varied processors, storage, and battery capacity.

\subsection{Layer Compression Technique Benchmark}
In our experiment, we study the performance differences of the state-of-the-art DNN compression techniques in terms of user demand metrics, \ie accuracy $A$, storage $S$, latency $T$, and energy cost $E$.
%
\lsc{For this benchmark, we use the default compression hyperparameters (\eg $k$ in both $W_{1f}$ and $W_2$) for a fair comparison.}


\subsubsection{Benchmark Settings}
\label{subsec:bench_set}
We apply ten mainstream compression techniques from three categories, \ie weight compression ($W_{1f}$, $W_{2}$, $W_{3}$, $W_{1c}$), convolution decomposition ($C_{1}$, $C_{2}$, $C_{3}$), and special architecture layers ($L_{1}$, $L_{2}$, $L_{3}$), \rev{to a 13-layer AlexNet (input, conv$_{1}$, pool$_{1}$, conv$_{2}$, pool$_{2}$, conv$_{3}$, conv$_{4}$, conv$_{5}$, pool$_{3}$, fc$_{1}$, fc$_{2}$, fc$_{3}$ and output)~\cite{model:alexnet} and compare their performance evaluated on CIFAR-10 dataset ($D_{2}$)~\cite{data:cifar}}
on a RedMi 3S smartphone.
The details of them are as follows. 
\begin{itemize}
\item \textbf{$W_{1f}$}: 
insert a fc layer between 
fc$_{i}$ and fc$_{(i+1)}$ layers using the singular value decomposition (SVD) based weight matrix factorization~\cite{bib:lane2016:IPSN}. 
The neuron number $k$ in the inserted layer is set as $k=m/12$, where $m$ is the number of neurons in fc$_{i}$.
\item \textbf{$W_{2}$}: 
insert a fc layer between
fc$_{i}$ and fc$_{(i+1)}$ using sparse-coding, another matrix factorization method ~\cite{bib:bhattacharya2016:CD-ROM}.
The $k$-basis dictionary used in $W_{2}$ is set as $k=m/6$, where $m$ is the neuron number in fc$_{i}$.
\item \textbf{$W_{3}$}: 
prune fc$_{1}$ and fc$_{2}$ \rev{using the magnitude based weight pruning strategy proposed in~\cite{bib:arXiv2015:han}.}
\rev{It removes unimportant weights whose magnitudes are below a threshold (\ie $0.001$).}
\item \textbf{$L_{3}$}: 
replace the fc layers, fc$_{i}$ and fc$_{i+1}$, with a global average pooling layer~\cite{bib:lin2013:NIN}.
\rev{It generates one feature map for each category in the last conv layer. 
The feature map is then fed into the softmax layer.}
\item \textbf{$W_{1c}$}: 
insert a conv layer between conv$_{i}$ and pool$_{i}$ using SVD based weight factorization~\cite{bib:lane2016:IPSN}.
The numbers of neurons $k$ in the inserted layer by SVD $k=m/12$, where $m$ is the neuron number in conv$_{i}$.
\item \textbf{$C_{1}$}: 
decompose conv$_{i}$ using convolution kernel sparse decomposition~\cite{bib:cvpr2015:ICCV}.
\rev{It replaces a conv layer using a two-stage decomposition based on principle component analysis.}
\item \textbf{$C_{2}$}: 
decompose conv$_{i}$ with depth-wise separable convolution~\cite{bib:arXiv2017:Howard}. 
The width multiplier $\alpha = 0.5$.
\item \textbf{$C_{3}$}: 
decompose conv$_{i}$ using the sparse random technique~\cite{bib:ICLR2017:soravit} and we set the sparsity coefficient $\theta = 0.75$.
\rev{The technique replaces the dense connections of a small number of channels with sparse connections between a large number of channels for convolutions. 
Different from $C_{2}$, it randomly applies dropout across spatial dimensions at conv layers.}
\item \textbf{$L_{1}$}: replace conv$_{i}$ by a Fire layer~\cite{bib:iandola2016:arxiv}.
\rev{A Fire layer is composed of a $1\times1$ conv layer and a conv layer with a mix of $1\times1$ and $3\times3$ conv filters.
It decreases the sizes of input channels and filters.}
\item \textbf{$L_{2}$}: replace conv$_{i}$ by a micro multi-layer \rev{perceptron} embedded with multiple small kernel conv layers (Mlpconv)~\cite{bib:lin2013:NIN}. 
\rev{It approximates a nonlinear function to enhance the abstraction of conv layers with small (\eg $1\times1$) conv filters.}
\end{itemize}
The parameters ($k$ in $W_{1f}$, $W_{1c}$ and $W_{2}$, the depth multiplier $\alpha$ in $C_{2}$, the sparse random multiplier $\theta$ in $C_{3}$) are empirically optimized by comparing the performance on the layer where the compression technique is applied.

\begin{figure}[t]
\centering
\includegraphics[width=0.48\textwidth]{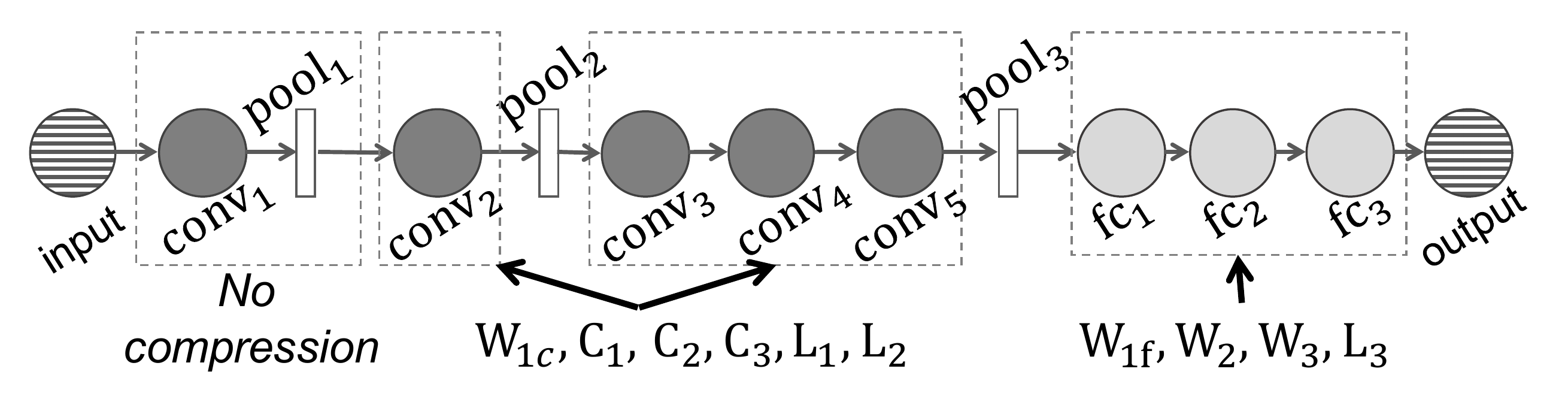}
\vspace{-2mm}
\caption{An illustration of the locations that different layer compression techniques are applied to AlexNet.}
\label{fig_baseline}
\vspace{-4mm}
\end{figure}

\begin{figure*}[ht]
  \centering
  \subfloat{
   \includegraphics[width=0.8\textwidth]{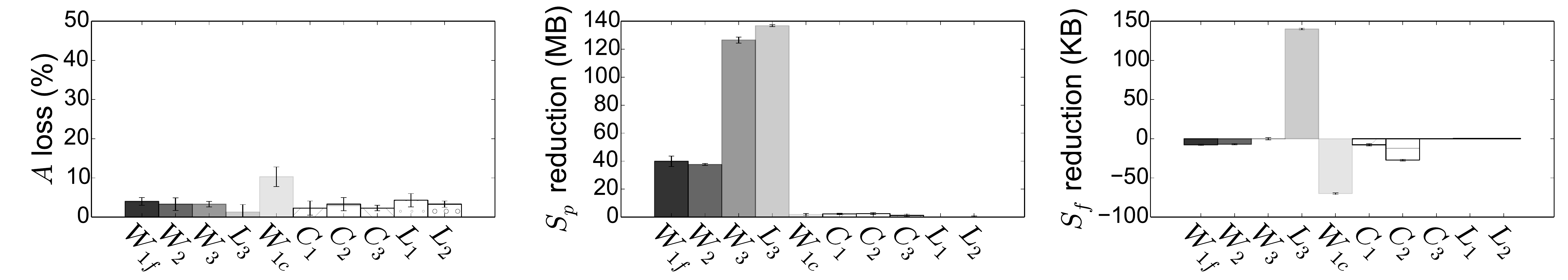}}
   \\
   \vspace{-3mm}
   \subfloat{
   \includegraphics[width=0.8\textwidth]{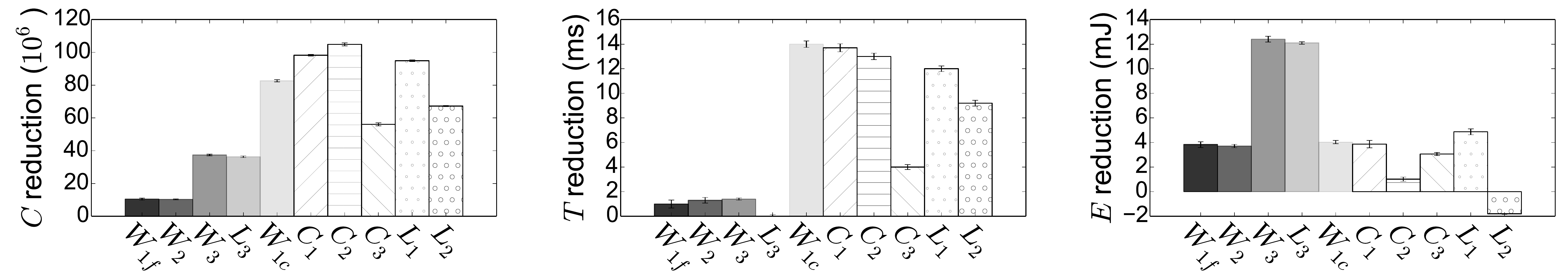}}
 \vspace{-4mm}
  \caption{Performance of different layer compression techniques minus by the \textit{initial layer} that they are applied to, in terms of accuracy A, storage ($S_p$, $S_f$), computational cost $C$, latency $T$, and energy cost $E$. 
The Y-axis denotes the accuracy loss ($\%$) over the initial \rev{AlexNet} and the cost reduction over the \textit{initial layer} that they are applied to.}
\vspace{-2mm}
  \label{fig_compare_compression}
\end{figure*}

As shown in \figref{fig_baseline}, compression techniques $W_{1f}$, $W_{2}$, $W_{3}$ and $L_{3}$ can be applied to the fc layers (fc$_1$, fc$_{2}$ and fc$_{3}$), while $W_{1c}$, $C_{1}$, $C_{2}$, $C_{3}$, $L_{1}$ and $L_{2}$ are employed to compress the conv layers (conv$_{2}$, conv$_3$, conv$_{4}$ and conv$_{5}$).
For each layer compression technique, we load the compressed DNN on smartphone to process the test data $10$ times, and obtain the mean and variance of the inference performance and resource cost, considering the varied workload of the device at different test times. 

\begin{figure*}[t]
  \centering
  \subfloat{
   \includegraphics[width=0.8\textwidth]{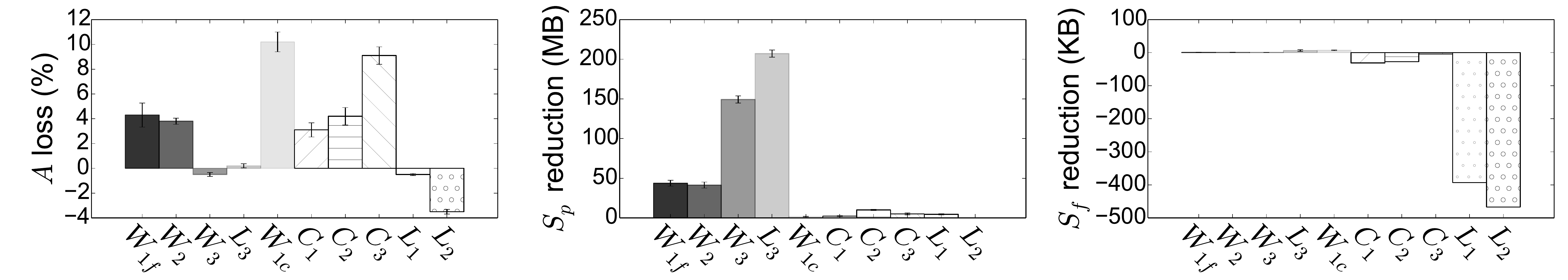}}
   \\
   \vspace{-3mm}
   \subfloat{
   \includegraphics[width=0.8\textwidth]{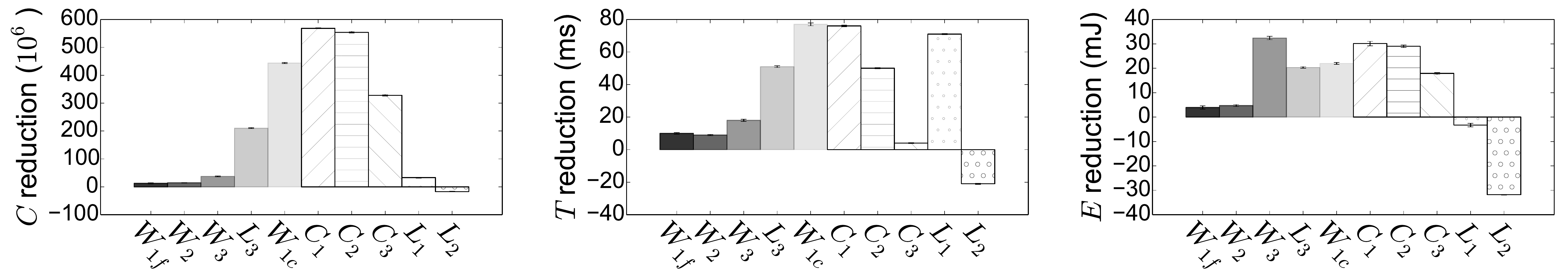}}
  \vspace{-4mm}
  \caption{Performance of different layer compression techniques normalized over the \textit{entire} AlexNet in terms of accuracy A, storage ($S_p$, $S_f$), computational cost $C$, latency $T$, and energy cost $E$. 
Y-axis denotes the accuracy loss ($\%$) and the cost reduction over the original \textit{entire} \rev{AlexNet}.}
\vspace{-1mm}
\label{fig_layer_to_model}
\end{figure*}

\subsubsection{Performance of Single Compression Technique}
\label{subsubsec:individual}
To illustrate the performance of different compression techniques, we compare their compressed DNNs in terms of the evaluation metrics ($A$, $S_p$, $S_f$, $T$ and $E$), over both the initial layer that they are applied to (see \figref{fig_compare_compression}) and the entire initial network, \ie \rev{AlexNet} (see \figref{fig_layer_to_model}). 
First, we can see that overall these mainstream compression techniques are quite effective in trimming down the complexity of the initial network, \rev{with a certain accuracy loss ($ 0.3 \% - 10.2 \%$) or accuracy gain ($0.5\% - 2.4\%$). 
For example, the compression techniques $W_{3}$ and $L_{3}$ reduce $S_p$ by about $150 - 203 MB$, while $W_{1c}$, $C_{1}$, $C_{2}$, $C_{3}$, $L_{1}$ and $L_{2}$ reduce $S_p$ to be less than $10MB$.}
Second, as expected, compressing the fc layers ($W_{1f}$, $W_2$, $W_3$, and $L_3$) results in a higher $S_p$ reduction, while compressing the conv layers ($W_{1c}$, $C_{1}$, $C_2$, $C_3$, $L_1$ or $L_2$) lead to a larger $C$ reduction. This is due to the common observation in DNNs that the conv layers consume dominant computational cost while the fc layers account for most of the storage cost.
\rev{
Third, most of the considered compression techniques affect the $S_f$ only in the order of $KB$, thus we only consider $S_p$ for the storage cost in following experiments.}

\textbf{Summary.}
The performance of different categories of compression techniques on the \textit{same} DNN varies.
Within the same category of compression techniques, the performance also differs. 
There is no a single compression technique that achieves the best $A$, $S$, $T$ and $E$. 
To achieve optimal overall performance on different mobile platforms and  applications, it is necessary to combine different compression techniques and tune the compression hyperparameters according to the specific usage demands.

\subsubsection{Performance of Blindly Combined Compression Techniques}
\label{subsec:manual_two}

\begin{table*}[h]
\scriptsize
\centering
\caption{\rev{Performance of combining two compression techniques to compress both the fc layers and the conv layers, evaluated on a RedMi 3S smartphone (Device 1) using the AlexNet model and CIFAR-10 dataset ($D_{1}$).} } 
\label{tb:combine}
\vspace{-3mm}
\begin{tabular}{|c|c|c|c|c|c|c|c|c|c|}
\hline
\multirow{2}{*}{\textbf{\begin{tabular}[c]{@{}c@{}}Compression \\ technique\end{tabular}}} & \multicolumn{4}{c|}{\textbf{ \rev{Measured accuracy \& cost} }} & \multirow{2} {*}{\textbf{\begin{tabular}[c]{@{}c@{}}Compression\\ technique\end{tabular}}} & \multicolumn{4}{c|}{\textbf{ \rev{Measured accuracy \& cost}  }} \\ \cline{2-5} \cline{7-10} 
 & $A(\%)$ & $S_p(MB)$ & $T(ms)$ & $E(mJ)$ &  & $A(\%)$ & $S_p(MB)$ & $T(ms)$ & $E(mJ)$ \\ \hline
\textbf{$C_{1}$+$W_{1f}$} & \rev{$74.2$} & \rev{$15.3$} & \rev{$180$} & \rev{$62.8$} & \textbf{$L_{1}$+$W_{1f}$} & \rev{$79.8$} & \rev{$16.2$} & \rev{$194$} & \rev{$33.7$} \\ \hline
\textbf{$C_{1}$+$W_{2}$} & \rev{$75.1$} & \rev{$12.3$} & \rev{$189 $} & \rev{$65.2$} & \textbf{$L_{1}$+$W_{2}$} & \rev{$78.1$} & \rev{$15.3$} & \rev{$189$} & \rev{$34.4$} \\ \hline
\textbf{$C_{1}$+$W_{3}$} & \rev{$77.6$} & \rev{$23.2$} & \rev{$132$} & \rev{$63.48$} & \textbf{$L_{1}$+$W_{3}$} & \rev{$84.8$} & \rev{$1.1$} & \rev{$86$} & \rev{$13.9$} \\ \hline
\textbf{$C_{1}$+$L_{3}$} & \rev{$75.4$} & \rev{$0.68$} & \rev{$102$} & \rev{$52.9$} & \textbf{$L_{1}$+$L_{3}$}\cite{bib:iandola2016:arxiv} & \rev{$87.1$} & \rev{$1.6$} & \rev{$257$} & \rev{$78.2$} \\ \hline
\textbf{$C_{2}$+$W_{1f}$} & \rev{$72.4$} & \rev{$15.3$} & \rev{$129$} & \rev{$33.1$} & \textbf{$L_{2}$+$W_{1f}$} & \rev{$86.4$} & \rev{$17.4$} & \rev{$305$} & \rev{$108.4$} \\ \hline
\textbf{$C_{2}$+$W_{2}$} & \rev{$71.8$} & \rev{$14.9$} & \rev{$130$} & \rev{$31.3$} & \textbf{$L_{2}$+$W_{2}$} & \rev{$86.9$} & \rev{$17.1$} & \rev{$312$} & \rev{$100.1$} \\ \hline
\textbf{$C_{2}$+$W_{3}$} & \rev{$81.8$} & \rev{$2.9$} & \rev{$124$} & \rev{$14.8$} & \textbf{$L_{2}$+$W_{3}$} & \rev{$88.7$} & \rev{$10.6$} & \rev{$266$} & \rev{$51.6$} \\ \hline
\textbf{$C_{2}$+$L_{3}$} & \rev{$81.5$} & \rev{$0.7$} & \rev{$98$} & \rev{$16.9$} & \textbf{$L_{2}$+$L_{3}$}\cite{bib:lin2013:NIN}& \rev{$87.1$} & \rev{$1.8$} & \rev{$126$} & \rev{$113.4$} \\ \hline
\textbf{$C_{3}$+$W_{1f}$} & \rev{$59.3$} & \rev{$16.7$} & \rev{$236$} & \rev{$43.4$} & \textbf{$W_{1c}$+$W_{1f}$} & \rev{$78.4$} & \rev{$16.1$} & \rev{$139$} & \rev{$36.1$} \\ \hline
\textbf{$C_{3}$+$W_{2}$} & \rev{$57.5$} & \rev{$15.7$} & \rev{$210$} & \rev{$42.7$} & \textbf{$W_{1c}$+$W_{2}$}\cite{bib:bhattacharya2016:CD-ROM} & \rev{$79.2$} & \rev{$16.4$} & \rev{$147$} & \rev{$39.3$} \\ \hline
\textbf{$C_{3}$+$W_{3}$} & \rev{$53.2$} & \rev{$3.2$} & \rev{$60$} & \rev{$21.7$} & \textbf{$W_{1c}$+$W_{3}$} & \rev{$61.2$} & \rev{$2.7$} & \rev{$143$} & \rev{$20.8$} \\ \hline
\textbf{$C_{3}$+$L_{3}$} & \rev{$77.3$} & \rev{$1.4$} & \rev{$84$} & \rev{$26.8$} & \textbf{$W_{1c}$+$L_{3}$} & \rev{$56.2$} & \rev{$1.2$} & \rev{$27$} & \rev{$22.9$} \\ \hline
\end{tabular}
\vspace{-4mm}
\end{table*}

In this experiment, we compare the performance when blindly combining two compression techniques, tested on a RedMi 3S smartphone (Device 1) using the AlexNet model and CIFA-10 dataset ($D_{1}$). 
Specifically, one of the four techniques to compress the fc layers fc$_{1}$ and fc$_{2}$ (\ie $W_f$, $W_{2}$, $W_{3}$ or $L_{3}$) is combined with one of the six techniques to compress the conv layer conv$_{2}$ (\ie $W_{1c}$, $C_{1}$, $C_{2}$, $C_{3}$, $L_{1}$ or $L_{2}$), leading to a total of 24 combinations. 
Among them, the $W_{1c}+W_{2}$, $L_1+L_3$ and $L_2 +L_3$ combinations have been introduced in the prior works named SparseSep \cite{bib:bhattacharya2016:CD-ROM}, SqueezeNet \cite{bib:iandola2016:arxiv} and NIN \cite{bib:lin2013:NIN}, respectively.

\tabref{tb:combine} summarizes the results. %
We leverage the compressed AlexNet using the $W_3$ technique as a baseline.
%
In particular, it achieves a detection accuracy of $79.9\%$ and requires a parameter storage of $6.09 MB$, an energy cost of $30.72 mJ$, and a detection latency of $189 ms$. 
First, compared with the compressed model using $W_3$, some combinations of  compression techniques, \eg $C_2$+$W_{3}$ and $C_{2}$+$L_{3}$, reduce more than $48 mJ$ of $E$, decrease $T$ by $103 ms$, and dramatically cut down $S_p$ by more than $18 MB$, while incurring only $2.4\%$ accuracy loss.
While some combinations might perform worse than a single compression technique, \eg $W_{1c}$+$L_{3}$ and $C_{3}$+$W_{3}$ incur over $28\%$ accuracy loss.  
Second, the combination of $L_{1}$+$W_{3}$ achieves the best balance between system performance and resource cost.

\textbf{Summary.}
Some combinations of two compression techniques can dramatically reduce the resource consumption of DNNs than using a single technique.
Others may lead to performance degradation.
Furthermore, the search space grows exponentially when combining more than two techniques.
These results demonstrate the need for an automatic optimizer to select and combine compression techniques.

\subsection{Performance of DRL Optimizer}
This section \lsc{tests the performance of the DDPG and DQN optimizer in hyperparameter search and compression technique selection, and evaluates the collaborative two  optimizers}.

\subsubsection{\lsc{Hyperparameters Learned by DDPG Optimizer}}
\label{subsec:hyperparameter}
\lsc{
We first describe the compression hyperparameters needed for our benchmark compression techniques, and present how we transform various hyperparameters to a "ratio" so that they can share a single DDPG agent with the same action space $As\sim [0,1]$.
%
As in \secref{subsec:bench_set}, we apply ten mainstream layer compression techniques at different conv and fc layers.
Note that only some of them need extra compression hyperparameters.
In particular, we consider the following "ratio" hyperparameters, whose optional value can be normalized as a percentage within the real-value region $[0,1]$: 
\begin{itemize}
\item ratio of the number of neurons inserted between $fc_i$ and $fc_{(i+1)}$ layer to the number of neurons at $fc_i$ by $W_{1f}$ technique. 
\item ratio of the number of neurons inserted between $conv_i$ and $conv_(i+1)$ layer to the number of neurons at $conv_i$ layer by $W_{1c}$ technique.
\item ratio of the number of k-basis dictionary inserted between $fc_i$ and $fc_(i+1)$ layer to the number of neurons at $fc_i$ layer by $W_2$.
\item ratio of the neuron number at layer $i$ used $W_3$ to neuron number in original DNN layer $i$.
\item width multiplier (a percentage) in $C_2$.
\item sparsity coefficient (a percentage) in $C_3$.
\end{itemize}
}

\begin{table}[t]
\scriptsize
\caption{The optimal hyperparameter at each compressed layer using different compression techniques, found by the proposed DDPG optimizer on [AlexNet, CIFAR-10(D2)].}
\label{tb_DDPG_hyper}
\vspace{-3mm}
\centering
\begin{tabular}{|c|c|c|c|c|c|c|}
\hline
\multirow{2}{*}{\textbf{Layer}} & \multicolumn{6}{c|}{\textbf{Hyperparameters of compression technique}} \\ \cline{2-7} 
 & \textbf{$W_{1f}$} & \textbf{$W_{1c}$} & \textbf{$W_2$} & \textbf{$W_3$} & \textbf{$C_2$} & \textbf{$C_3$} \\ \hline
\textbf{conv1} & - & - & - & - & - & - \\ \hline
\textbf{conv2} & - & 0.39 & 0.21 & 0.89 & 0.44 & 0.99 \\ \hline
\textbf{conv3} & - & 0.73 & 0.25 & 0.78 & 0.64 & 0.28 \\ \hline
\textbf{conv4} & - & 0.98 & 0.24 & 0.82 & 0.98 & 0.97 \\ \hline
\textbf{conv5} & - & 0.93 & 0.22 & 0.89 & 0.82 & 0.01 \\ \hline
\textbf{fc1} & 0.74 & - & 0.91 & 0.86 & - & - \\ \hline
\textbf{fc2} & 0.19 & - & 0.89 & 0.81 & - & - \\ \hline
\textbf{fc3} & - & - & - & - & - & - \\ \hline
\end{tabular}
\end{table}
\vspace{-1mm}

\lsc{
\tabref{tb_DDPG_hyper} presents the performance of the DDPG optimizer on hyperparameter search and provides a referential hyperparameter setup in the compressed AlexNet CIFAR-10 (D2) using different layer compression techniques.
The first conv layer and final fc layer are not compressed.
$W_{1f}$, $W_{1c}$ and $W_2$ conduct weight factorization at conv and fc layers using an inserted layer with $19\%$ to $74\%$ neurons.
$W_3$ prunes the weights of both conv and fc layers by the compression ratio of $21\%$ to $89\%$.
$C_2$ and $C_3$ decompose conv layers by the sparsity multiplier ranging from $1\%$ to $99\%$.
}

\lsc{
\textbf{Summary.} 
The optimal hyperparameters of the single compression technique at different layers differ. The search space is large when searching the optimal hyperparameters for multiple layers. 
To balance the compression performance and the searching cost, an automated layer-wise hyperparameter search optimizer is necessary.
}

\subsubsection{Performance Comparison of Optimizer}
\label{subsubsec:performance_optimizer}
This experiment is to evaluate the advantage of both the proposed \textit{DQN optimizer} and \textit{DDPG optimizer} when searching for the optimal compression combination as well as hyperparameters. 
To do so, we compress [LeNet, MNIST] and [AlexNet, CIFAR-10] using the DQN optimizer, the two-phase DRL optimizer and two baseline optimization schemes and evaluate the resulted DNNs on a RedMi 3S snartphone (Device 1). The accuracy loss ($\% $) and the cost reduction ($\times$) are normalized over the compressed DNNs using the $W_3$ technique.

\begin{itemize}
 \item \textbf{Exhaustive optimizer}: This scheme exhaustively test the performance of all combinations of two compression techniques (similar to \secref{subsec:manual_two}), and select the best trade-off on the validation dataset of MNIST, \ie the one that yields the largest reward value defined by Eq.~\eqref{eq_multiplier}. The selected one is $L_{2}$+$L_{3}$, \ie \textit{Fixed}, \rev{in both the cases of LeNet on MNIST and AlexNet on CIFAR-10}.
\lsc{The selected combination $L_2 +L_3$ does not have tunable hyperparameters.}
 
 \item \textbf{Greedy optimizer}:
It loads the DNN layer by layer and selects the compression technique that has the largest reward value defined by Eq.~\eqref{eq_multiplier}, in which both $\mu_1$ and $\mu_2$ are set to be 0.5. Also,
when $T$ or $S$ violate the budget $T_{bgt}$ or $S_{bgt}$, the optimization terminates.
\lsc{The compression hyperparameters layer compression techniques are fixed by the default optimal value (similar to \secref{subsec:bench_set}).}
\item \textbf{DQN optimizer}: It compresses the DNN using the DQN optimizer as described in \secref{sec:optimizer}. 
We set the scaling coefficients in Eq.~\eqref{equ_DRL_R2} to be $\mu_1=0.6$ and $\mu_2=0.4$ considering that the battery capacity in RedMi 3S is relatively large and thus the energy consumption is of lower priority, and we set $\mu_3=0.5$ and $\mu_4=0.5$ in Eq.~\eqref{equ_DRL_R2} because their corresponding constraints (\ie $C$ and $S_{p}$) are equally important. The same as in the Greedy search within this subsection.
\lsc{The compression hyperparameters of layer-wise compression techniques are also set as the default optimal value (similar to \secref{subsec:bench_set}).}

\item \lsc{\textbf{DDPG plus DQN optimizer}}: It further leverages the DDPG optimizer to tune the compression hyperparameters of the  DNN compressed by above DQN optimizer. The setup of scaling coefficients ($\mu_1=0.6, \mu_2=0.4, \mu_3=0.5, \mu_4=0.5 $) is the same as that in the DQN optimizer within this subsection.
\end{itemize}

\begin{table*}[]
\scriptsize
\centering
\caption{Performance of the best DNN generated by the DQN, \lsc{the two-phase DRL optimizer (\ie DDPG plus DQN)}, and two baseline optimizers are tested on a RedMi 3S smartphone using LeNet on MNIST (case 1) and AlexNet on CIFAR-10 (case 2). The accuracy loss $\% $ and the cost reduction ($\times $) are normalized over the corresponding DNN compressed using $W_3$.}
\label{tb_optimizer_compare}
\vspace{-3mm}
\begin{tabular}{|c|c|c|c|c|c|c|c|c|}
\hline
\multirow{2}{*}{\textbf{Optimizer}} & \multicolumn{4}{c|}{\textbf{Compared to the \rev{compressed LeNet on MNIST (case 1)} }} & \multicolumn{4}{c|}{\textbf{Compared to the \rev{compressed AlexNet on CIFAR-10 (case 2)} }}  \\ \cline{2-9}
 & $A$ loss & $S_p$ & $T$ & $E$ & $A$ loss & $S_p$ & $T$ & $E$   \\ \hline
\textbf{Exhaustive} & \rev{$0.1\%$} & \rev{$23.9\times$} & \rev{$2.7\times$} & \rev{$1.1\times $} & \rev{ $-7.2\%$} & \rev{$3.5\times$} & \rev{$0.7\times$} & \rev{$1.2\times$}  \\ \hline
\textbf{Greedy} & \rev{$2.3\%$} & \rev{$4.6\times$} & \rev{$0.6\times$} & \rev{$2.7\times$} & \rev{$0.3\%$} & \rev{$2.2\times$} & \rev{$1.2\times$} & \rev{$ 1.9\times$} \\ \hline
\textbf{DQN} & \rev{$0.4\%$} & \rev{$24.5\times$} & \rev{$3.1\times$} & \rev{$2.4\times $} & \rev{$ -2.6\% $} & \rev{$2.5\times $} & \rev{$2.5\times $} & \rev{$ 1.4\times$}  \\ \hline
\textbf{\lsc{DDPG plus DQN}} & \rev{$0.2\%$} & \rev{$28.5\times$} & \rev{$3.8\times$} & \rev{$2.8\times $} & \rev{$ -4.9\% $} & \rev{$4.6\times $} & \rev{$2.3\times $} & \rev{$ 1.8\times$}  \\ \hline
\end{tabular}
\vspace{-3mm}
\end{table*}

\tabref{tb_optimizer_compare} summarizes the best performance achieved by the above four optimizers. 
We can see that the networks generated by  DQN and DDPG optimizer achieve better overall performance in terms of storage $S_p$, latency $T$, and energy consumption $E$, while incurring negligible accuracy $A$ loss ($0.1\%$ or $2.1\%$), compared to those generated by the other two baseline optimizers. 
In particular, compared with the DNN compressed by $W_3$, the best DNN from the Greedy optimizer only reduces $S_p$ by $4.6\times$ and $2.2\times$ in [LeNet, MNIST] (case 1) and [AlexNet, CIFAR-10] (case 2), respectively.
In contrast, the best DNN from the Exhaustive optimizer, \ie \textit{Fixed}, can reduces $S_p$ by $23.9\times$ and $3.5\times$, respectively. \lsc{DQN optimizer cuts down $24.5\times$ and $2.5 \times$ of $S_p$}, while \lsc{DDPG plus DQN} optimizer achieves a maximum reduction of $28.5\times$ and $4.6\times$ on $S_p$ in  two cases.
Second, the network from the \lsc{proposed DQN and DDPG plus DQN optimizers are the most effective in reducing the latency ($> 2.3\times$) in both cases, while those from the two baseline optimizers may result in an increased $T$ in some cases. 
For example, the DDPG plus DQN optimizer reaches the maximum reduction of $T$ by $3.8\times$ in case 1, and the DQN optimizer sharply reduces $T$ by $2.5\times$ in case 2.
}
The network from the Greedy optimizer increases $T$ by $0.6\times$ in case 1 and the one from the Exhaustive optimizer introduces an $0.7\times$ extra $T$ in case 2.
Third, when comparing the energy cost $E$, \textit{Fixed} is the least energy-efficient (reduce $E$ by only $1.1\times$ over the DNN compressed by $W_3$), while those from \lsc{the DQN, the DDPG plus DQN, and the Greedy optimizers} achieve an reduction of $1.4\times$ to $2.8\times$, respectively.
Meanwhile, the accuracy loss from the two baseline optimizers ranges from $0.1\%$ to $2.3\%$, while those from DQN plus DDPG optimizer achieves the best accuracy (only a $0.2\%$ degradation in case a and even a $4.9\%$ gain in case 2). 
Finally, as for the training time, the \lsc{DDPG and DQN} optimizers require a shorter, or equal, or longer time compared with the exhaustive and Greedy optimizers (refer to $\S$ 7.4.2).
%

\textbf{Summary.}
\lsc{The proposed DDPG and DQN optimizers attain the best overall performance in both experiments.
Both DDPG plus DQN and DQN optimizers} outperform the other two schemes for DNN compression in terms of the storage size, latency, and energy consumption while incurring negligible accuracy in diverse recognition tasks. 
This is because the run-time performance metrics ($A$, $S$, $T$ and $E$) and the resource cost ($S$ and $T$) \lsc{of the whole DNN network} are systematically included in the reward value and adaptively feedback to the \lsc{layer-wise compression technique selection or hyperparameter search process}.


\subsection{Performance of AdaDeep}
In this subsection, we test the end-to-end performance of \systemname over six tasks and on twelve mobile platforms.
\rev{Furthermore, to show the flexibility of AdaDeep in adjusting the optimization objectives based on the user demand, we show some examples of the choices on the scaling coefficients in Eq.~\eqref{equ_DRL_R2}.}

\subsubsection{AdaDeep over Different Tasks}
In this experiment, \systemname is evaluated on all the six tasks/datasets  using a RedMi 3S smartphone (Device 1).
\rev{We set the scaling coefficients in Eq.~\eqref{equ_DRL_R2} to be the same as those for the DRL optimizer in $\S$ 5.3.1, \ie $\mu_1=0.6$ and $\mu_2=0.4$, $\mu_3=0.5$ and $\mu_4=0.5$}. In addition, we assume a Cache storage budget of $2$ MB and a latency budget of $10$ ms.

\begin{table*}[ht]
\scriptsize
\centering
\caption{Performance of \systemname evaluated on different datasets using a RedMi 3S smartphone (Device 1), normalized over the corresponding \rev{DNNs compressed using $W_3$}. \textbf{The compression techniques marked by `*' are the novel combinations that have not been proposed in related studies}.}
\label{tb:tasks}
\vspace{-4mm}
\begin{tabular}{|c|c|c|c|c|c|c|}
\hline
\multirow{2}{*}{Task} & \multirow{2}{*}{Compression techniques$ \& $hyperparameters}  & \multicolumn{5}{c|}{Compare to the DNN compressed by $W_3$} \\ \cline{3-7} 
 &  & $S_{p}$ & $C$ & $T$ & $E$ & $A$ loss \\ \hline
1.MNIST (LeNet)  & $*C_3(0.96,0.24)+W_3(0.85)$ & \rev{$ 1.8 \times$} & \rev{$1.5\times $} & \rev{$ 1.8\times$} & \rev{$1.3\times$} &  \rev{$-2.5 \%$} \\ \hline
\rev{2.CIFAR-10 (AlexNet)} & \rev{${L_1+W_3(0.78,0.82)}$} & \rev{$4.6\times$}  &  \rev{$3.1\times $}  & \rev{$2.3\times$} & \rev{$1.8\times$} &  \rev{$-4.9 \%$} \\ \hline
\lscrev{3.CIFAR-100 (ResNet-56)} & \lscrev{${L_1+W_3(0.48,0.52)}$} & \lscrev{$1.7\times$}  &  \lscrev{$1.9\times $}  & \lscrev{$1.1\times$} & \lscrev{$1.2\times$} &  \lscrev{$-0.1 \%$} \\ \hline
4.ImageNet (AlexNet) & $* L_2+C_2(0.88,0.81)+L_3$ & \rev{$18.5\times$} & \rev{$2.3\times$} & \rev{$3.6 \times$} & \rev{$1.4\times$} & \rev{$-1.2\%$} \\ \hline
5.ImageNet (VGG) & $* L_2+C_1+L_3$ & \rev{$37.3\times $} & \rev{$2.3\times$} & \rev{$18.6\times$} & \rev{$4.1\times$} & \rev{$0.2\%$} \\ \hline
6.Ubisound (LeNet) & $* C_3(0.83, 0.31)+L_3$ & \rev{$3.2\times$} & \rev{$\rev1.9\times$} & \rev{$1.6\times$} & \rev{$1.1\times$} &  \rev{$0.4\%$} \\ \hline
7.Har (LeNet) & $L_1+W_3(0.76)$ & \rev{$2.1\times$} & \rev{$0.8\times$} & \rev{$0.8\times$} & \rev{$1.5\times$} & \rev{$-2.6\%$} \\ \hline
\end{tabular}
\vspace{-4mm}
\end{table*}

\textbf{Performance.} 
\tabref{tb:tasks} compares the performance of the best DNNs generated by \systemname on the \lscrev{six} tasks in terms of accuracy loss, storage $S_p$, computation $C$ (total number of MACs), latency $T$ and energy cost $E$, normalized over the DNNs \rev{compressed using $W_3$}. 
Compared with their initial DNNs, DNNs generated by \systemname can achieve a reduction of \lscrev{$1.7\times$ -  $37.3\times$ in $S_{p}$, $0.8\times$ -  $3.1\times$ in $C$, $0.8\times$ -  $18.6\times$ in $T$, and $1.1\times$ -  $4.3\times$ in $E$, with a negligible accuracy loss ($< 0.4\%$) or even accuracy gain ($< 4.9\%$).}

\textbf{Summary.} 
For different compressed DNNs, tasks, and datasets, the combination of compression techniques found by \systemname also differs.
Specifically, the combination that achieves the best performance while satisfying the resource constraints is $C_{3}$+$W_{3}$ for Task 1 (on MNIST initialized using LeNet), \lscrev{$L_{1}$+$W_{3}$ for Task 2 (on CIFAR-10 initialized using AlexNet),} $L_{1}$+$W_{3}$ for Task 3 (on CIFAR-100 initialized using ResNet-56), $L_{2}$+$C_{2}$+$L_{3}$ for Task 4 (on ImageNet initialized using AlexNet), $L_{2}$+$C_{1}$+$L_{3}$ for Task 5 (on ImageNet initialized using VGG), $C_{3}$+$L_{3}$ for Task 6 (on Ubisound initialized using LeNet), and $L_{1}$+$W_{3}$ for Task 7 (on Har initialized using LeNet), respectively.
\rev{We can see that although the combination of compression techniques found by AdaDeep cannot always outperforms a single compression techniquein in all metrics, it achieves a better overall performance in terms of the five metrics according to the specific user demands.}

\subsubsection{AdaDeep over Different Mobile Devices}
This experiment evaluates \systemname across twelve different mobile devices using LeNet and UbiSound ($D_{4}$) as the initial DNN and evaluation dataset, respectively. The performance achieved by the initial DNN is as follows: $A = 95.1\%$, $S_p = 25.2$ MB, $C = 28,324,864$, $T = 31$ ms, and $E=4.3$ mJ. 

Different devices have different resource constraints, which lead to different performance and budget demands and thus require different coefficients $\mu_1 \sim \mu_4$ in Eq.~\eqref{equ_DRL_R2}. Specifically, we empirically optimize $\mu1 \sim \mu4$ for different devices to be:
$\mu_2=max\{ \frac{4000-E_{battery}}{4000}, 0.6\}$, $\mu_1=1-\mu_2$,
$\mu_4=max\{\frac{8-S_{Cache}}{8}, 0.6\}$, and $\mu_3=1-\mu_4$.

\begin{table*}[t]
\scriptsize
\centering
\caption{Performance of \systemname on different devices using the UbiSound dataset ($D_{4}$), normalized over the corresponding initial DNNs. \textbf{The compression techniques marked by `*' are the combinations that have not been proposed in related studies}.}
\label{tb:devices}
\vspace{-3mm}
\begin{tabular}{|c|c|c|c|c|c|c|c|}
\hline
\multirow{2}{*}{\begin{tabular}[c]{@{}c@{}}Device\end{tabular}} & \multirow{2}{*}{Compression techniques $\&$ hyperparameters}  & \multicolumn{5}{c|}{Compare to initial DNN} \\ \cline{3-7} 
 &  & $S_{p}$ & $C$ & $T$ & $E$ & $A$ loss \\ \hline
1. Xiaomi Redmi 3S & $C_{1}$+$W_{3}(0.81)$ & $12.1\times$ & $2.1\times$ & $1.6\times$ & $1.1\times$ &  0.9 \% \\ \hline
2. Xiaomi Mi 5S & $C_{2}(0.41,0.48,0.77,0.65)$+$L_{3}$ & $27.1\times$ & $3.6\times$ & $2.1\times$ & $1.2\times$ &  1.8 \% \\ \hline
3. Xiaomi Mi 6 & \textbf{$*C_{2}(0.65,0.68,0.97,0.65)$+$W_{3}(0.83)$} & $13.1\times$ & $5.6\times$ & $1.9\times$ & $1.6\times$ & 1.1\% \\ \hline
4. Huawei pra-al00 & \textbf{$*C_{2}(0.63,0.66,0.96,0.85)$+$W_{3}(0.81)$} & $12.7\times$ & $6.8\times$ & $1.4\times$ & $1.8\times$ & 1.2\% \\ \hline
5. Samsung note5 & \textbf{$*C_{2}(0.63,0.68,0.94,0.83)$+$W_{3}(0.81)$} & $12.8\times$ & $4.1\times$ & $1.6\times$ & $1.8\times$ &  1.2\% \\ \hline
6. \rev{Huawei} iP9 & \textbf{$C_{1}$+$W_{3}(0.82)$} & $13.0\times$ & $1.6\times$ & $1.6\times$ & $1.7\times$ & 0.9\% \\ \hline
7. Sony watch S$W_{3}$ & $C_{2}(0.73,0.86,0.98,0.86)$+$W_{2}(0.89)$ & $6.4\times$ & $2.1\times$ & $1.5\times$ & $9.8\times$ &  1.6\% \\ \hline
8. Huawei watchH2P  & $L_{2}$+$L_{3}$ & $27.8\times$ & $3.6\times$ & $3.1\times$ & $8.3\times$ &  2.1\% \\ \hline
9. firefly-rk3999 & \textbf{$L_{1}$+$W_{3}(0.83)$} & $13.2\times$ & $5.6\times$ & $2.6\times$ & $1.2\times$ & 1.8\% \\ \hline
10. firefly-rk3288 & $C_{2}(0.63,0.68,0.97,0.85)$+$W_{1f}(0.21)$ & $3.4\times$ & $4.8\times$ & $1.1\times$ & $1.3\times$ & 0.7\% \\ \hline
11. Xiaomi box 3S & \textbf{$*C_{3}(0.89,0.48,0.95,0.12)$+$W_{3}(0.84)$} & $14.1\times$ & $4.1\times$ & $1.4\times$ & $1.1\times$ &  1.2\% \\ \hline
12. Huawei box & $L_{1}$+$L_{3}$ & $28.1\times$ & $1.6\times$ & $2.8\times$ & $1.2\times$ & 1.9\% \\ \hline
\end{tabular}
\vspace{-3mm}
\end{table*}

\textbf{Performance.} 
\tabref{tb:devices} summarizes the generated compression combinations \lsc{as well as compression hyperparameters} by \systemname and the corresponding preformance. 
For twelve different resource constraints, DNNs generated by \systemname, which are initiated with the same DNN model, can reduce parameter size by $3.4\times$ - $28.1\times$, computation cost by $1.6\times$ - $6.8\times$, latency by $1.1\times$ - $3.1\times$ and energy cost by $1.1\times$ - $9.8\times$, respectively, while incurring a negligible accuracy loss ($\leq$ 2.1\%). 
The optimal combinations of compression techniques found by \systemname differ from device to device. Furthermore, \systemname finds some combinations that work the best for a given mobile platform yet have not been proposed by previous works (\eg $C_{1}$+$W_{3}$ for Device 1, $C_{2}$+$W_{3}$ for Devices 3, 4 and 5, $C_{3}$+$W_{3}$ for Device 11).

\rev{The training process of \systemname includes three intertwined phases: training the regularized DNN, re-training (such as in $L_1$, $L_2$, and $L_3$) or fine-tuning (such as in $W_3$) DNN for compression, and training the DRL (\ie DQN and DDPG) optimizer. Because the training time of the regularized DNN is standard, we only quantify the total training time required by the DNN compression and the DRL based selection on different tasks, which is ~$3$ hours on [MNNIST, LeNet], $~10$ hours on [CIFAR-10, LeNet], $~6.5$ hours on [CIFAR-10, AlexNet], $~16$ hours on [CIFAR-100, ResNet], $~3.5$ hours on [Ubisound, LeNet], $~2$ hours on [Har, LeNet], and $~15$ hours on [ImageNet, AlexNet], respectively, using two HP Z400 workstations with two GEFORCE GTX 1060 GPU cards.}

\textbf{Summary.}
Overall, \systemname can automatically \rev{select} the proper combinations of compression techniques that meet diverse demands on accuracy and resource constraints within $3.5$ to $15$ hours. We find that the optimal compression strategy differs over tasks and across mobile devices, and there is no one-fit-all compression technique for all tasks and mobile devices.
\systemname is able to adaptively select the best compression strategy given diverse user demands. It also uncovers some combinations of compression techniques not proposed in previous works.
\rev{Also, the sensitivity of the performance metrics to different resources may vary for different choices of the scaling coefficients ($\mu_1 \sim \mu_4$).}

\postsec
\presec
\section{Related Work}
\label{sec:related}
Our work is closely related to the following research.
\vspace{-2mm}
\subsection{Automatic Hyperparameter Optimization}


Hyperparameters of DNNs, such as the number of layers and neurons, the size of filters and the model architecture, are crucial to the inference accuracy. 
Common hyperparameter tuning techniques can be categorized into parallel search, such as grid search~\cite{bib:bergstra2011:ANIPS} and random search~\cite{bib:bergstra2012:JMLR}, and sequential search, \eg Bayesian optimization~\cite{bib:snoek2012:ANIPS}.
The grid and random search approaches search blindly and thus are usually time-consuming.
Bayesian approaches~\cite{bib:snoek2015:ICML} \cite{bib:domhan2015:IJCAI} \cite{bib:springenberg2016:ANIPS} automatically optimize hyperparameters, but is slow due to the sequential operations.

Inspired by state-of-the-art automatic hyperparameter optimization techniques, compressing DNNs can be viewed as a hyperparameter tuning process. 
\systemname is the first to treat compression technique as a tunable coarse-grained hyperparameter. And it provides a systematic method to automatically search the most suitable coarse-grained hyperparameter (\ie compression technique) and the fine-grained compression hyperparameters.


\vspace{-2mm}
\subsection{DNN Compression}
The success of machine learning in mobile and IoT applications has stimulated the adoption of more powerful DNNs in mobile and embedded devices \cite{bib:UbiComp15:Lane, bib:TNET17:Zheng, ISCAS_PredictiveNet, NIPS-18, DeepkMeans}.
Compression is a commonly employed technique to trim down the complexity of DNNs, which can be performed by reducing the weight precision, or the number of operations, or both~\cite{bib:sze2017:arxiv}.
Various DNN compression techniques have been proposed, including weight compression~\cite{bib:bhattacharya2016:CD-ROM} \cite{bib:arXiv2015:han} \cite{bib:lane2016:IPSN}, convolution decomposition~\cite{bib:ICLR2017:soravit} \cite{bib:arXiv2017:Howard} \cite{bib:cvpr2015:ICCV}, and compact architectures~\cite{bib:iandola2016:arxiv} \cite{bib:lin2013:NIN}.
However, existing efforts investigate a one-for-all scheme, \eg reducing DNN complexity using one compression technique, and do not consider the diversity of user demands on performance and resource cost.
Our experiment results show that there is no single compression technique work well for diverse user demands.

\systemname enables an automatic selection of the best combination of compression techniques to balance the application-driven system performance and the platform-imposed resource constraints. 
Specifically, \systemname supports automatic selection from three categories of mainstream DNN compression techniques, and automatic configuration of compression hyperparameters. 
 
\vspace{-2mm}
\subsection{Run-time DNN Optimization}
Orthogonal to DNN compression, DNNs can also be optimized at run-time to reduce their resource utilization and unnecessary overhead on energy, latency, storage or computation.
MCDNN~\cite{bib:mobisys2016:han} pre-evaluates a set of compressed models with different execution cost and selects one for each DNN that maximizes the accuracy given total cost constrains of multi-programmed DNNs. 
However, it only presents two cost reduction algorithms. 
LEO~\cite{bib:georgiev2016:mobicom} designs a low power unit resource scheduler to maximize energy efficiency for the unique workload of different tasks on heterogenous computation resources.
DeepX~\cite{bib:lane2016:IPSN} designs a set of resource control algorithms to decompose DNNs into different unit-blocks for efficient execution on heterogeneous computation resources.
EIE~\cite{bib:han2016:press} is a dedicated accelerator to execute sparse NN.
 
The above run-time optimization techniques can be applied on top of the compressed DNN generated by \systemname to further improve the efficiency of DNN execution on mobile devices.
For example, the current version of \systemname only leverages the CPU on mobile platforms for DNN execution.
The scheduler proposed in~\cite{bib:georgiev2016:mobicom} and ~\cite{bib:lane2016:IPSN} can be combined when extending \systemname to mobile platforms with heterogeneous resources.
With proper hardware support, the sparse NN output by \systemname can also be executed faster using the accelerator in~\cite{bib:han2016:press}.

\vspace{-2mm}
\subsection{Automatic Control Techniques using DRL}
Deep reinforcement learning (DRL) is widely applied in automatic-play games to learn actions at different states that maximize a given reward function~\cite{bib:mnih2013:arxiv}.
For example, Mnih~\etal~\cite{bib:mnih2013:arxiv} propose to learn control policies from complex sensory inputs using a deep Q-network (DQN).
Liu~\etal~\cite{bib:acc2017:liu} leverage DQN to dynamically select parts of a NN to execute according to different input resolution so as to improve computational efficiency of multi-objective optimization problems. 
Achiam~\etal~\cite{bib:achiam2017:ICML} solve the constrained optimization problem with DRL by replacing the objective and constraints with approximate surrogate, \ie lower bound on policy divergence.
However, the required operation of inverting the divergence matrix is in general impractically expensive.
Bello~\etal~\cite{bib:bello2017:arvix} present a framework to tackle the combinatorial optimization of sequential problems with DRL and recurrent DNN.
\lsc{David~\etal~\cite{bib:ICLM2014:silver} apply deterministic policy gradient to choose action from continuous action space}.

To the best of our knowledge, \systemname is the first work to leverage DQN and DDPG for DNN compression technique selection as well as compression hyperparameter optimization, considering both application-driven system performance and platform constraints. 

\vspace{-2mm}
\subsection{Automated DNN Architecture Optimization}
An emerging topic for the deep learning community is to automate the engineering process of deep model architectures: using recurrent networks and reinforcement learning to generate the model descriptions of deep models~\cite{zoph2016neural}, or by transferring architectural building blocks to construct scalable architectures on larger datasets \cite{zoph2017learning}. 
Those methods are purely data-driven, with deep architectures composed with the goal to maximize the expected accuracy on a validation set. 
Lately, a handful of exploratory works have emerged to correlate the model composition with domain knowledge. %
For example, Andreas~\etal~\cite{andreas2015deep} constructed and learned modular networks, which composed collections of jointly-trained neural "modules" into deep networks for question answering, to simultaneously exploit the representational capacity of deep networks and the compositional linguistic structure of questions. 
Devin~\etal~\cite{devin2017learning} proposed a similar modular network by decomposing robotic policies into task-specific and robot-specific modules, to facilitate multi-task and multi-robot policy transfer. 
However, none of those previous efforts have correlated their efforts with DNN compression and energy efficiency. 

\vspace{-2mm}
\subsection{AutoML for DNN Compression}
Automated machine learning (AutoML) aims at providing effective system to free non-experts from selecting the right algorithm or hyperparameter at hand.
AutoML systems like Auto-WEKA~\cite{bib:kotthoff2017:MLR} and Auto-skelearn~\cite{bib:feurer2015:ANIPS} leverage Bayesian optimization method to search the best classifier given the datasets.
Auto-Net~\cite{bib:mendoza2016:AML} leverages the tree-based Bayesian method to tune DNN hyperparameters without human intervention.
AMC~\cite{bib:eccv2018:he} comes up with a continuous compression ratio control strategy with  DDPG agent to find the redundancy.
AdaNet~\cite{bib:ICML2017:cortes} adaptively learn both the DNN structure and its weights.
NetAdapt~\cite{bib:eccv2018:he} is an automatic tool to gradually reduce the number of filters of a DNN for resource consumption reduction.

So far AutoML systems do not yet simultaneously support DNN architecture and hyperparameter optimization for DNN compression.
\systemname extends the automation of DNN architecture selection and hyperparameter optimization to include DNN compression, that considerably trading off among both user-defined requirements and platform-imposed constraints.

\postsec
\presec
\section{Discussions}
\label{sec:discuss}
\lscrev{
In this section, we point out several limitations of \systemname in this work for future research.}

\zimu{
\fakeparagraph{Finer-grained Compression Optimization}
\systemname is built upon a set of predefined compression techniques.
It searches for the best combination of compression techniques and the corresponding hyperparameters for each layer via reinforcement learning.
Hence the optimization space is constrained by the granularity of the predefined compression techniques.
Integration of other categories of compression techniques will expand the action space and potentially result in better compressed DNNs.
Randomization techniques such as layer skipping or re-ordering may also facilitate finer-grained model compression optimization.
}


\zimu{
\fakeparagraph{Extensions to Other Layer Types and Processors}
In this work, we mainly design \systemname based on compression techniques for dense and convolutional layers. 
Since compression techniques for recurrent layers are gaining increasing attention \cite{RNN}, nne next step is to extend our optimization framework to also support recurrent layers.
Furthermore, although our evaluations include experiments with twelve different mobile devices, we mainly evaluate the performance of different methods on devices with merely CPUs.
With the increasing popularity of GPUs installed on off-the-shelf mobile devices, it remains an interesting question how \systemname performs on those mobile devices.
}



\zimu{
\fakeparagraph{DRL Optimization Speedup}
Despite its effectiveness in optimizing the compression technique combination and hyperparameters, DRL can consume considerable time due to the large search space and the sophisticated optimization procedure. 
We anticipate that conditional search by setting search conditions based on activation and previous prediction at different layers/compression techniques will accelerate the optimization process.
}



\postsec
\presec
\section{Conclusion}
\label{sec:conclusion}
This paper presents \systemname, \lscrev{a usage-driven and automated DNN compression and optimization framework that selects the most suitable combination of compression techniques and the corresponding compression hyperparameters to} balance diverse user-specified performance goals and device-imposed resource constraints.
\lscrev{
We systematically formulate user demands on performance requirements (\eg accuracy, latency) and resource constraints (\eg storage and energy budgets) into a unified optimization problem.
And we leverage two types of DRL optimizors, \ie a DQN based optimizer and a DDPG based optimizer, to effectively find the feasible combination of compression techniques and the corresponding compression hyperparameters in a layer-wise manner.
%
Evaluations on six widely used tasks and twelve different devices show that there is no one-fit-all compression technique or hyperparameter configuration that meets the diverse user demands.}
%
\systemname also figures out some novel combinations of compression techniques unexplored in previous work.
\lscrev{
\systemname is the first to model DNN compression as 
an automated hyperparameter tuning process, that automates the selection of the coarse-grained hyperparameters (\eg compression techniques) and the fine-grained compression hyperparameters (\eg compression ratio and sparsity coefficient) of DNNs.}
 

\postsec

\section*{Acknowledgements}
We are grateful for Professor Lin Zhong (Rice University) for his useful feedback on an early version of this paper, Xin Wang, Yuheng Wei, and Bo Deng (Xidian University) for their help on implementing some of the baseline techniques.
This work is supported in part by National Key Research $\&$ Development Program of China ($\#2018YFB1003605$), Natural Science Foundation of China, NSFC ($\#61472312$), Open Fund of State Key Laboratory of Computer Architecture ($\#CARCH201704$), the Youth Innovation Team of Shaanxi Universities, Shaanxi Found ($\#2018JM6125$, $\#B018230008$), and Natural Science Foundation (NSF) Award ($\#1801865$).


\bibliographystyle{IEEEtran}
\bibliography{reference}

\begin{IEEEbiography}[{\includegraphics[width=1in,height=1.25in,clip,keepaspectratio]{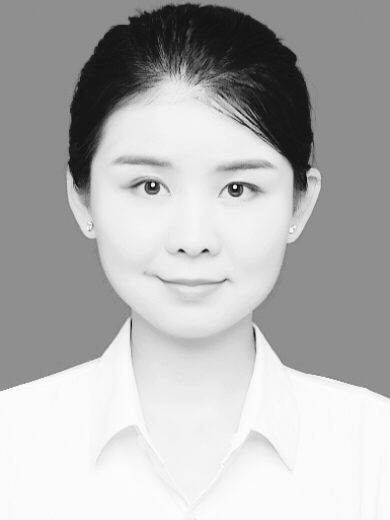}}]{Sicong Liu}
received the BS and MS degrees from Xidian University in 2013 and 2016, respectively. From 2017 to 2018, she was a visiting scholar at Rice University. She is currently a Ph.D. student with the School of Computer Science and Technology, Xidian University. Her research interests include mobile computing system, mobile and embedded deep learning design, and automated deep model optimization.
\end{IEEEbiography}

\begin{IEEEbiography}[{\includegraphics[width=1in,height=1.25in,clip,keepaspectratio]{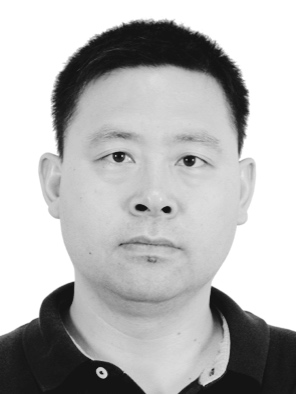}}]{Junzhao Du}
received the BS, MS, and Ph.D. degrees from School of Computer Science at Xidian University in 1997, 2000, and 2008, respectively. He is currently a professor and PhD advisor at Xidian University. His research interests include mobile computing, cloud computing, and IoT systems. He is the member of ACM/IEEE, senior member of CCF, and vice secretary of ACM Xi’an Chapter.
\end{IEEEbiography}

\begin{IEEEbiography}[{\includegraphics[width=1in,height=1.25in,clip,keepaspectratio]{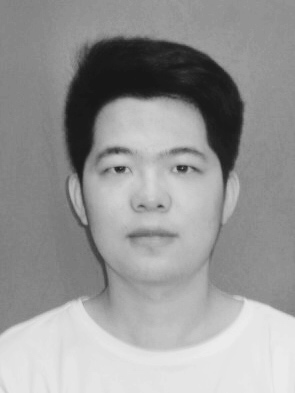}}]{Kaiming Nan}
received the BS degree in School of Software from Xidian University in 2017. He is currently a MS student at Xidian University. His research interests include mobile computing, mobile and embedded data collection, and energy consumption optimization in mobile deep learning.
\end{IEEEbiography}

\begin{IEEEbiography}[{\includegraphics[width=1in,height=1.25in,clip,keepaspectratio]{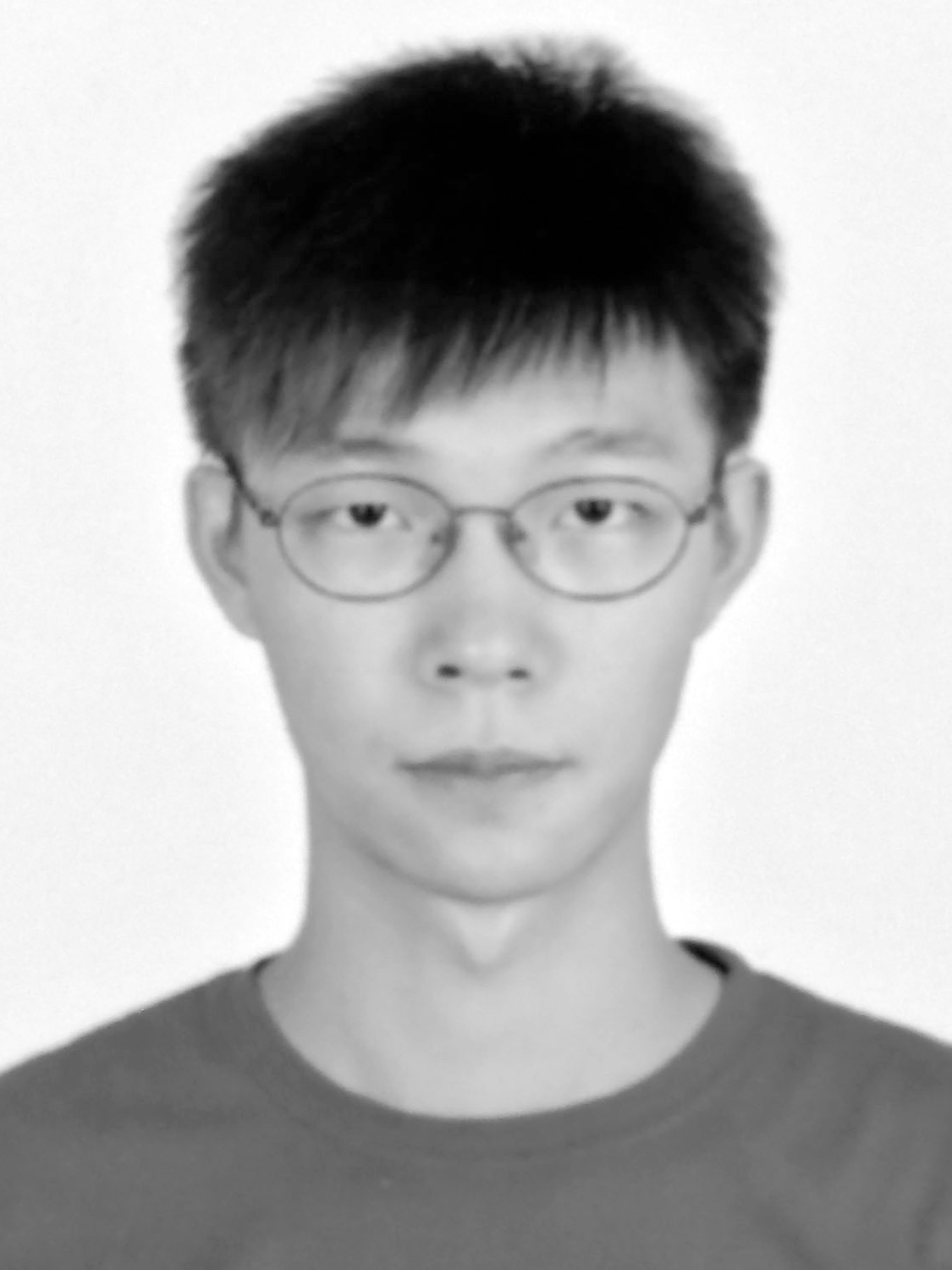}}]{Zimu Zhou}
is currently an Assistant Professor of Information Systems at Singapore Management University. He received the B.E. degree from the Department of Electronic Engineering, Tsinghua University, Beijing, China, in 2011, and the Ph.D. degree from the Department of Computer Science and Engineering, Hong Kong University of Science and Technology, Hong Kong, in 2015. From 2016 to 2019, he was a Post-Doctoral Researcher at the Computer Engineering and Networks Laboratory, ETH Zurich, Zurich, Switzerland. His research interests include mobile and ubiquitous computing. 
\end{IEEEbiography}

\begin{IEEEbiography}[{\includegraphics[width=1in,height=1.25in,clip,keepaspectratio]{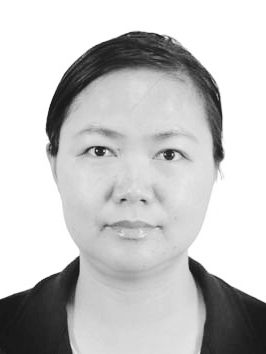}}]{Hui Liu}
received the BS, MS, and PhD degrees from School of Computer Science at Xidian University in 1998, 2003, and 2011, respectively. She is currently an associate professor at Xidian University. Her research interests includes big data analysis, task scheduling, and mobile computing. She is the member of ACC, IEEE, and CCF.
\end{IEEEbiography}

\begin{IEEEbiography}[{\includegraphics[width=1in,height=1.25in,clip,keepaspectratio]{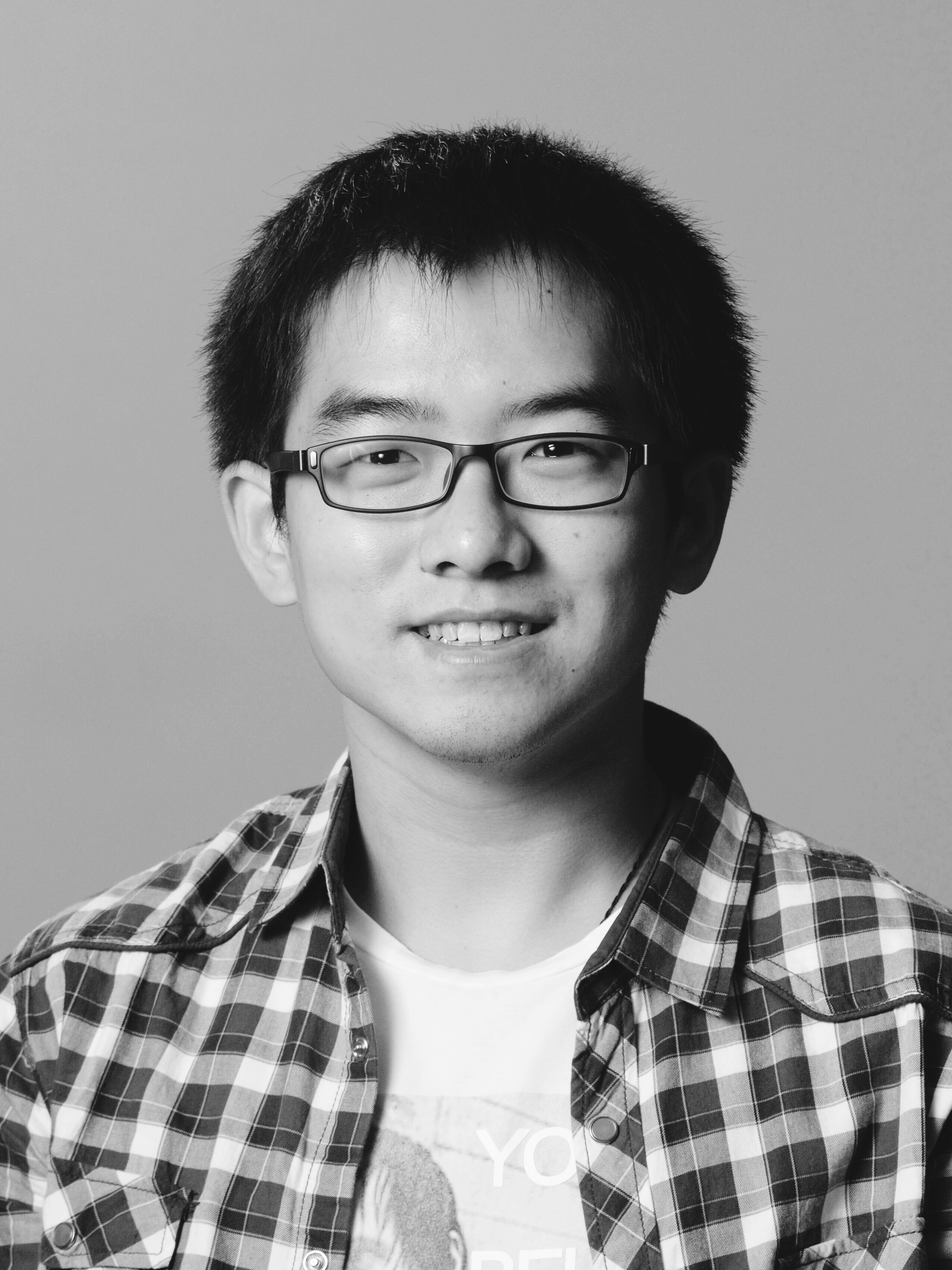}}]{Zhangyang (Atlas) Wang}
is an Assistant Professor of Computer Science and Engineering (CSE), at the Texas A$\&$M University (TAMU). During 2012-2016, he was a Ph.D. student in the Electrical and Computer Engineering (ECE) Department, at the University of Illinois at Urbana-Champaign (UIUC). Prior to that, he obtained the B.E. degree at the University of Science and Technology of China (USTC), in 2012. He was a former research intern with Microsoft Research (summer 2015), Adobe Research (summer 2014), and US Army Research Lab (summer 2013). Dr. Wang's research has been addressing machine learning, computer vision, as well as their interdisciplinary applications, using advanced feature learning and optimization techniques. He has co-authored over 80 papers, and has published 2 books and 1 invited chapter. He has been granted 3 patents, and has received over 20 research awards and scholarships.
\end{IEEEbiography}

\begin{IEEEbiography}[{\includegraphics[width=1in,height=1.25in,clip,keepaspectratio]{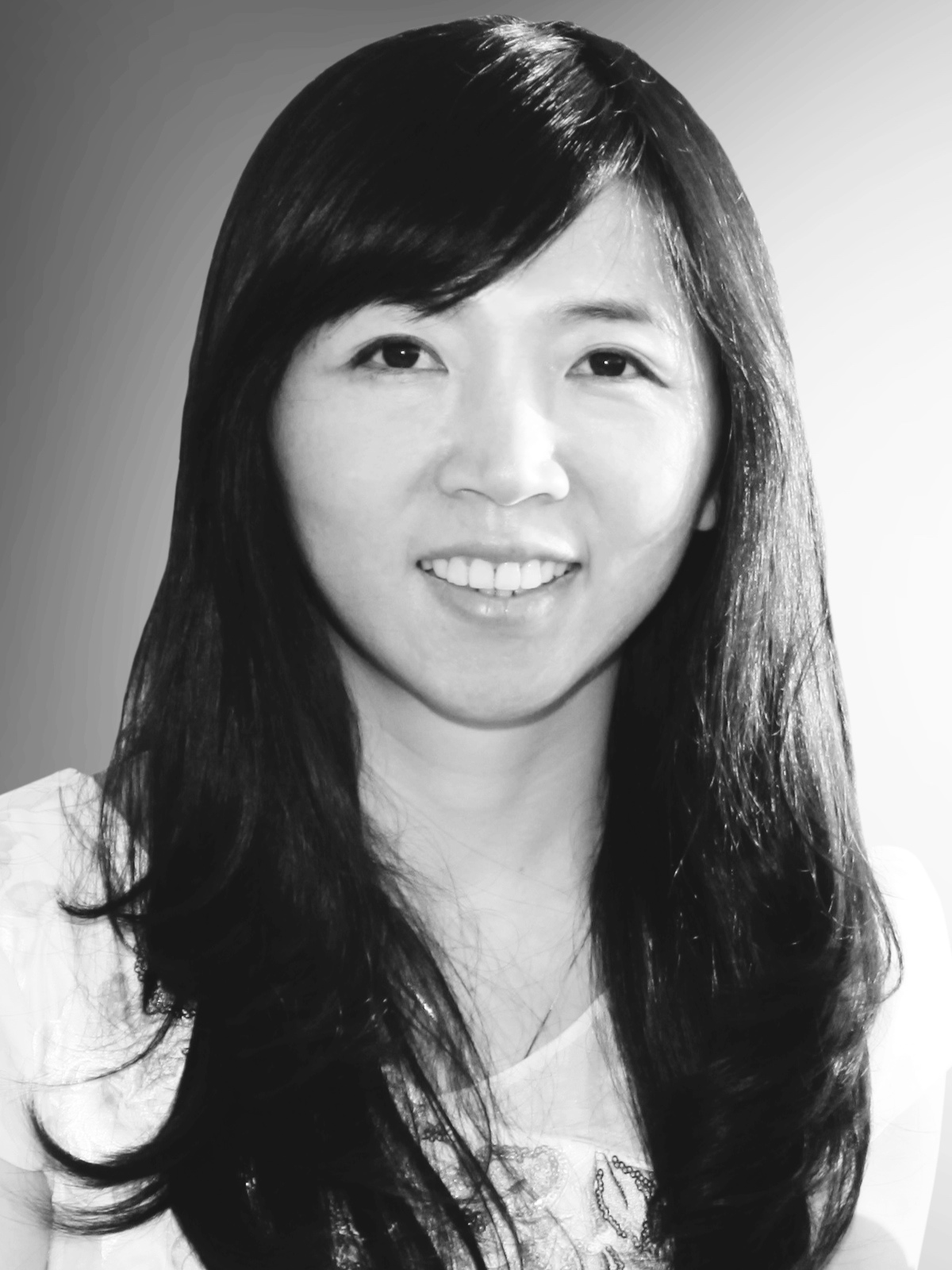}}]{Yingyan Lin}
is an Assistant Professor in the Department of Electrical and Computer Engineering (ECE) at Rice University. She received a Ph.D. degree in ECE from the University of Illinois at Urbana-Champaign in 2017. From 2007 to 2009, she worked at China's National Research Center for Integrated Circuits in Wuhan, where she designed three analog and mixed-signal circuit IPs for large panel displays that were acquired by TOSHIBA Microelectronics Corporation in Japan. She was the recipient of a Best Student Paper Award at the 2016 IEEE International Workshop on Signal Processing Systems (SiPS 2016), the 2016 Robert T. Chien Memorial Award at UIUC for Excellence in Research, and was selected as a Rising Star in EECS by the 2017 Academic Career Workshop for Women at Stanford University. Her research focuses on embedded machine learning, which is to explore algorithm-, architecture-, and circuit-level techniques for enabling powerful yet power hungry machine learning systems to be deployed in resource-constrained platforms.
\end{IEEEbiography}

\end{document}